\documentclass[10pt,twocolumn,letterpaper]{article}

\usepackage{cvpr} % To produce the REVIEW version
\usepackage[dvipsnames]{xcolor}
\usepackage{dblfloatfix}
\usepackage{afterpage}
\newcommand{\ourmethod}{MeshUp}
\newcommand{\ourtechnique}{BSD}

\definecolor{cvprblue}{rgb}{0.21,0.49,0.74}
\usepackage[pagebackref,breaklinks,colorlinks,citecolor=cvprblue]{hyperref}
\usepackage[percent]{overpic}

\title{\ourmethod{}: Multi-Target Mesh Deformation via Blended Score Distillation}

\author{Hyunwoo Kim\\
University of Chicago\\
\and
Itai Lang\\
University of Chicago\\
\and 
Noam Aigerman\\
University of Montreal\\
\and
Thibault Groueix\\
Adobe Research\\
\and 
Vladimir G. Kim\\
Adobe Research\\
\and
Rana Hanocka\\
University of Chicago\\
}

\begin{document}
\twocolumn[{
\begin{center}
    \vspace{-2cm}
    \maketitle
    \includegraphics[width=\linewidth]{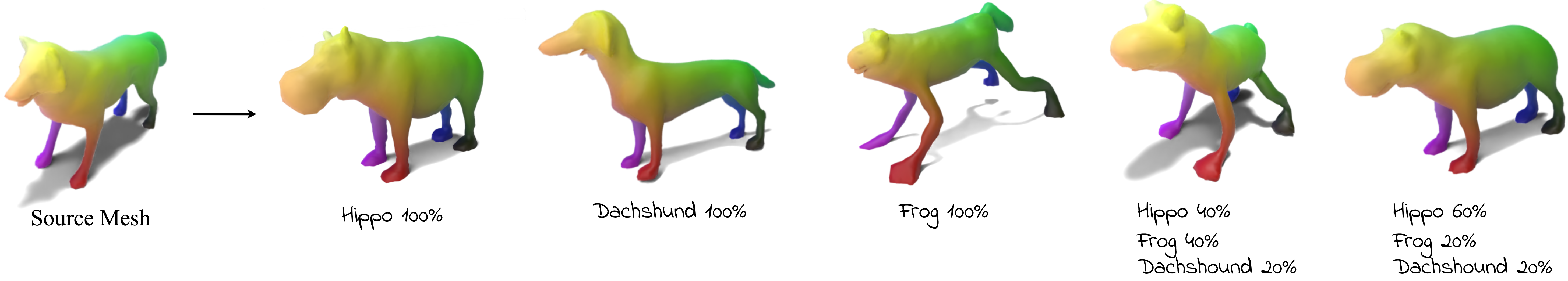}
    \captionof{figure}{\ourmethod{} is capable of deforming a source mesh into various concepts and into their weighted blends. The target objectives can be text prompts, images, or even mesh. Users can also input a set of \textit{control vertices} to explicitly define where on the mesh the particular concepts should be expressed (Figure ~\ref{fig:local_def1}). The colors on the mesh visualize the point-wise correspondence between the source and the deformed mesh.}
    \label{fig:teaser}
\end{center}%
}]
\begin{abstract}
We propose \ourmethod{}, a technique that deforms a 3D mesh towards multiple target concepts, and intuitively controls the region of their expression. Conveniently, the concepts can be defined as either text queries, e.g., ``a dog'' and ``a turtle,'' or inspirational images, and the local regions can be selected as any number of vertices on the mesh. We can effectively control the influence of the concepts and mix them together using a novel score distillation approach, referred to as the Blended Score Distillation (\ourtechnique{}). \ourtechnique{} operates on each attention layer of the denoising U-Net of a diffusion model as it extracts and injects the per-objective activations into a unified denoising pipeline from which the deformation gradients are calculated. To localize the expression of these activations, we create a probabilistic Region of Interest (ROI) map on the surface of the mesh, and turn it into 3D-consistent masks that we use to control the expression of these activations. We demonstrate the effectiveness of \ourtechnique{} empirically and show that it can deform various meshes towards multiple objectives. Our project page is at \href{https://threedle.github.io/meshup/}{https://threedle.github.io/meshup/}.
\end{abstract}

\section{Introduction}
\label{Introduction}

Deforming mesh is a central task in geometry processing~\cite{sorkine2004laplacian, mitra2007dynamic,sorkine2007rigid,solomon2011killing,corman2019functional,dodik2023variational}. In particular, it maintains valuable predefined attributes, such as artist-generated tessellation, UV map, textures, and motion functions.
Deforming a mesh, however, still remains a task that requires significant expertise, making it difficult for non-experts to creatively manipulate 3D models without knowing their low-level attributes. 
Addressing this challenge requires an intuitive, high-level control over 3D shapes in a way that can induce any non-expert users' creative workflows. In this work, we explore the use of diffusion to enable a user-friendly deformation-based 3D content generation.

In addition to the ease of use, creative workflows in generative tasks are also inspired from their ability to synthesize novel imagery--namely, by combining a range of diverse concepts~\cite{thecuriouscaseofuncurious}. Some cognitive theories even suggest that the ability to synthesize novel combinations of known concepts and exploring these conceptual ideas is essential to human creativity~\cite{philofcreativity}. 
While most methods that achieve 3D content generation optimize an implicit representation defined over 3D space~\cite{poole2022dreamfusion, lin2023magic3d, gaussiansplatting}, these representations are often inappropriate for mesh-specific tasks and cannot reuse any of the attributes defined over an artist-generated mesh. On the other hand, deformation-based approaches such as~\cite{Gao_2023_SIGGRAPH} lack the tools to enable a high-level, creative workflow for users to create novel conceptual imagery (e.g., "a creature with a bear's head and a frog's legs", or mixing across multiple targets) and achieve precise control over their expressions.

Motivated by this observation, we propose \ourmethod{}, a novel approach that deforms a source mesh towards multiple target concepts defined using a variety of inputs (texts, images, and even meshes), and localizes the region where these concepts are manifested. 
Given as input various types of user-defined ``concepts,'' their respective weights, and optionally a set of vertex points on the mesh, our method deforms a mesh to appropriately conform to a localized, weighted mixture of these concepts. 

\begin{figure}[tb!]
\centering
%trim={<left> <lower> <right> <upper>}
\includegraphics[width=\linewidth]{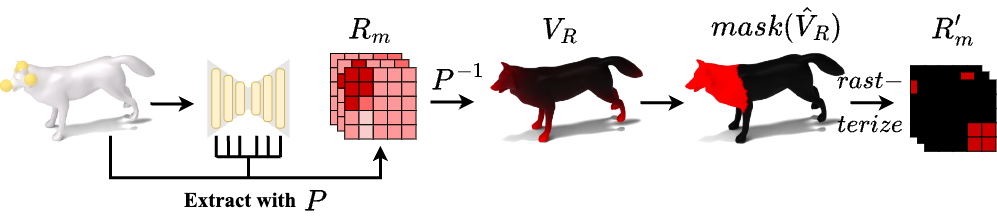} 
\vspace{-8mm}
\captionof{figure}{\textbf{Overview of Mask Extraction for Localized Control.}
For Localized control, we first extract the self-attention maps that correspond to each \textit{control vertex}. Using the inverse vertex-to-pixel map, $P^{-1}$, we then project the self-attention maps onto the mesh vertices and create a 3D attention map that describes the Region of Interest (ROI) per target concept ($V_R)$. We normalize and threshold this ROI map to create a 3D mask ($mask(\hat{v}_R))$, and rasterize the map from the same viewpoints as the mesh renderings to generate $R'_m$, the localization masks to be applied to the cross attention layers of the BSD pipeline.
}
\label{fig:system_local}
\end{figure}

In order to create a mixture of various concepts, we blend the activation maps by running the denoising process for each target and injecting the corresponding maps into a unified denoising U-Net, a method we call Blended Score Distillation (BSD).
We then estimate the gradients from the diffusion using Score Distillation Sampling (SDS), a method that enables the inference step of a diffusion model to be performed in a stochastic manner ~\cite{poole2022dreamfusion, scorejacobian}, and optimize the mesh deformation parameters, which we represent as Jacobians~\cite{aigerman2022neural}.
For fine-grained control over user-specified local regions, our framework additionally takes as input a set of selected vertices, each for a corresponding concept. Then for these concepts, we create a probability map over the mesh surface by extracting the self-attention maps from a diffusion process run on a batch of multi-viewpoint renderings, and reversely mapping them back to the surface. We then rasterize this probability map to create an attention mask that we use to control the region of deformation within our \ourtechnique{} pipeline (see Figure~\ref{fig:system_local}).

We leverage this novel pipeline to build a comprehensive creative modeling tool for concept mixing. The key features of our tool are (1) the support for mesh deformation towards multiple targets, (2) the capability to control both the strength and the region of their expression, (3) the ability to use either text, images, or other meshes as inputs.

\section{Related Work}
\noindent
\textbf{Image Editing Using Diffusion.}
Following the success of text-to-image generative Diffusion Models~\cite{DDIM, DDPM, latentdiffusion,  safelatent, GLIGEN, T2I, selfattention}, many diffusion-based image editing models \cite{semanticguidance, zhang2023adding, BLIP, DiffEdit, instructpix2pix, paintbyexample, hertz2022prompttoprompt, nulltextinversion, choi2023customedit, deltadenoising} have been developed. These methods allow introducing custom concepts~\cite{ruiz2023dreambooth,gal2022textual}, or enable fine-grained control of which regions and aspects of the image change~\cite{hertz2022prompttoprompt,instructpix2pix, DiffEdit} by weighting, modifying, and transferring the attention weights and activation of the diffusion networks.

\begin{figure}[tb!]
\centering
\includegraphics[width=\columnwidth]{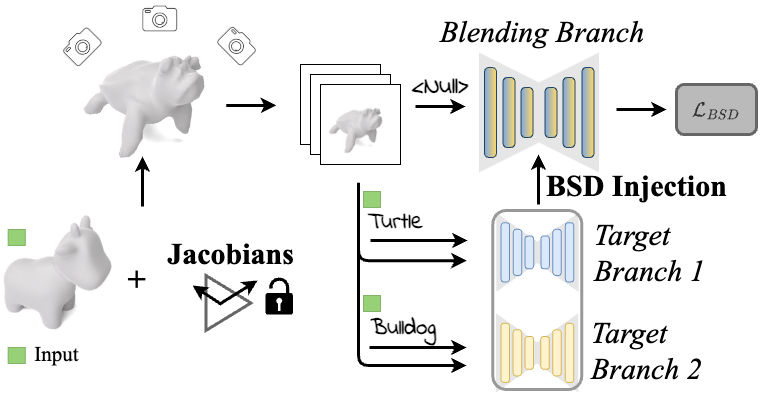} 
\captionof{figure}{{\bf Overview of Concept Blending.}  \ourmethod{} takes as input a 3D mesh and several target objectives, such as the text ``Sea turtle'' and ``Bulldog.'' We deform the source mesh by optimizing the per-triangle Jacobians of the mesh.
 At each iteration, we render the mesh and apply the same random noise for each target objective. Then, we pass the noised renderings and the text input through the U-Net of a pretrained text-to-image model and store the activations associated with each objective (the \textit{Target Branch)}. In the \textit{Blending Branch}, we feed the noised rendering of the mesh to the U-Net, but condition it on the null-text embedding. We blend and inject the activations stored at each target branch into the blending branch. The gradients from the blending branch are then backpropagated via Score Distillation Sampling (SDS). After running this process iteratively, the mesh is deformed into a blend of ``Sea turtle'' and ``Bulldog.''
}
\label{fig:system}
\end{figure}

\vspace{3mm}
\noindent
\textbf{Text-to-3D.}
These pretrained 2D diffusion generative techniques have also been used to enable 3D generation. This is usually accomplished by optimizing a 3D representation so that its rendering matches the desired text prompt~\cite{poole2022dreamfusion, makeit3d, prolificdreamer, lin2023magic3d, sjc, seo2023let, sella2023voxe, reimagine, highfidelity, efficientdreamer, sweetdreamer, gaussiansplatting, progressive3d, zero123, zhang2023text2nerf, text2mesh, shi2024mvdreammultiviewdiffusion3d, decatur20233dpaintbrushlocalstylization}.
These methods often rely on implicit fields as a 3D representation (e.g., NeRFs~\cite{mildenhall2020nerf}), which limits their editability, and often requires additional mesh conversion to support standard graphics pipelines. While some techniques allow editing these implicit fields~\cite{sella2023voxe,nerfediting, sine, controlnerf}, it is harder to provide local surface control, preserve correspondences (or use them to define continuous interpolations), with these models. A mesh can be extracted as a post-process~\cite{poole2022dreamfusion} using marching cubes~\cite{LorensenC87} and even further fine-tuned to match the desired prompt~\cite{lin2023magic3d, wang2023prolificdreamer}, but these meshes would not be consistent with one another, and automatic methods do not produce artist-quality tessellations or UV mappings, necessary for a production-ready asset. 
In this work, we instead use deformation of a single reference shape guided by multiple concepts (e.g,. textual prompts), which enables retaining necessary characteristics of the artist-created asset and enables to create a continuous semantic space interpolating between the concepts.

\vspace{3mm}
\noindent
\textbf{Mesh Deformation.}
Traditional mesh deformation is typically based on optimization of correspondences between vertices, faces, or other predictors that derive from these properties. \cite{sorkine2004laplacian,sorkine2007rigid} use energy minimizing functions to give users control over the deformation space, while~\cite{skinningcourse:2014,Fulton:LSD:2018} use skinning-based methods that interpolates the coordinate space with respect to the user handles. ~\cite{10.1145/2766963} uses optimal transportation to approximate correspondence across shapes. ARAP~\cite{sorkine2007rigid} and Laplacian surface editing~\cite{sorkine2004laplacian} use a variational formulation to regularize the deformation in a way that preserves details and prevents drifting of the geometry. However, these methods do not contain any semantics in their deformation and do not perform concept mixing.

Deforming a template mesh to various concepts has been explored even before the advances in neural networks~\cite{controlhandle}, but these techniques required user annotations for rigging meshes via handles and assigning semantic labels. 

Several data-driven techniques have been used to learn deformations~\cite{aigerman2022neural, hanocka2018alignet, yifan2020neural, Maesumi23, Groueix19, Uy21, eisenberger2021neuromorph, li20214dcompletenonrigidmotionestimation}. 
A recent class of works leverages text prompts as user inputs for driving a deformation towards an arbitrary textual prompt~\cite{text2mesh,CLIPMesh,Gao_2023_SIGGRAPH,fusiondeformer}. These methods use various deformation representations, and we opt to leverage Jacobians since they produce smooth and large-scale global deformations. 

We also observe that the CLIP objective lacks a full understanding of object details and that Diffusion-based objective, such as SDS~\cite{poole2022dreamfusion} provides better guidance. 
The main goal of this work is to extend these techniques to multi-target deformation, and provide tools to mix, edit, and explore the space of concept combinations.

\section{Method} \label{sec:method}

The primary goal of our method is, given $N$ sets of texts or image inputs that define the target "concepts," and their associated weights, $w$, to deform a source mesh into a shape that represents an effective mixture of these concepts, and control the "strength" of their expression using the weights. To that end, our method runs multiple diffusion pipelines in parallel and mixes their activation matrices within a unified pipeline to yield a single gradient direction that respects the appropriate weighted mixture of the target concepts.

For a framework that deforms a mesh into a specific target, two major design choices should be considered: the objective function (loss) and the representation of the mesh (the parameters to be optimized). For the objective function, we choose the Score Distillation Sampling (SDS) approach~\cite{poole2022dreamfusion, sjc}, a prevalent generative technique that allows the diffusion inference process to be performed in a stochastic manner and thus enables our deformation process to be performed with viewpoint-consistency. While a straightforward application of this objective to mesh deformation would be to directly optimize the vertex positions, this method often leads to sharp artifacts~\cite{CLIPMesh} or restricts deformations to only local adjustments~\cite{text2mesh}. On the other hand, Jacobian-based deformation has been proposed for smooth, continuous, and global deformations, but it has only been used with an $L_2$ supervision~\cite{aigerman2022neural} and CLIP similarity loss~\cite{Gao_2023_SIGGRAPH}. Using the SDS objective to supervise the optimization of the Jacobians offers a robust deformation framework with a powerful diffusion-based objective. 

In this section, we first overview how one might approach a single target deformation using a combination of Jacobian-based mesh deformation and SDS guidance. We then extend this concept to multi-target deformation via our novel Blended Score Distillation and explain how we achieve local control over the deformations.

\medskip
\noindent
\textbf{Jacobian-Based Mesh Deformation.} Our mesh deformation is represented by a per-face Jacobian matrix $J_i\in\mathbf{R}^{3\times 3}$, where the deformation of a mesh (vertex positions) is computed by optimizing the following least squares problem (\ie, Poisson Equation):
\begin{equation}
    \gamma^*={min}_{\gamma} \sum_i t_i ||\nabla_i(\gamma) - J_i||_2^2, 
    \label{eq:poisson}
\end{equation}

\noindent where $\gamma^*$ is the deformation map embedding the mesh such that its Jacobians $\nabla_i(\gamma)$ are as close as possible to the target Jacobians $\{J_i\}$, the parameters we optimize, and $\{t_i\}$ are the triangle areas. Similar to previous works, we use a differentiable Poisson solver layer~\cite{aigerman2022neural} to compute the deformation map, and a differentiable renderer~\cite{nvdiffrast} to connect this representation to image-based losses~\cite{Gao_2023_SIGGRAPH}. 

\begin{figure}
\centering
\includegraphics[clip, width=\linewidth]{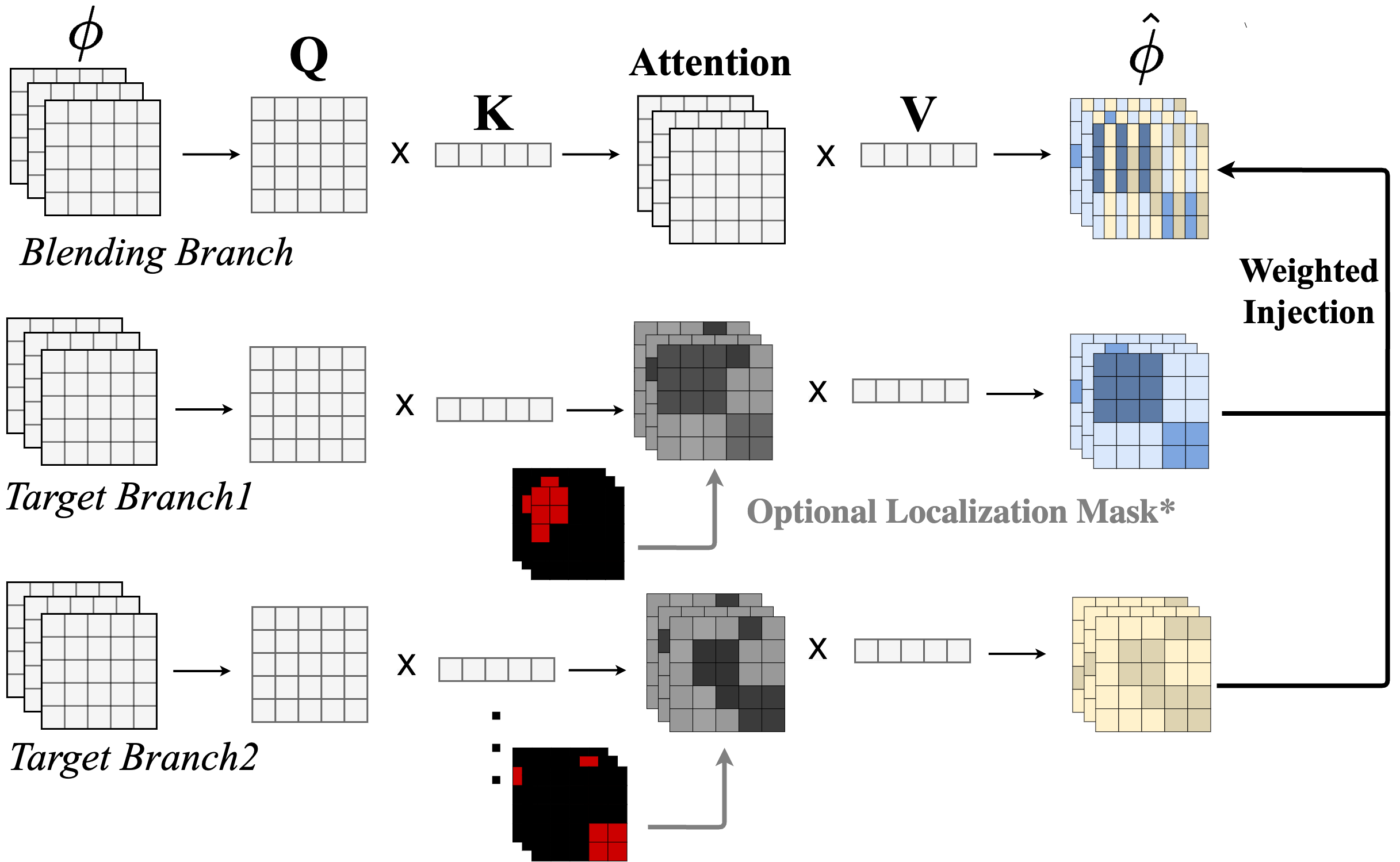} 
\captionof{figure}{{\bf  Overview of Blended Score Distillation (BSD).} For each attention layer in the denoising U-Net, we inject the activation maps from \textit{Target Branch1} and \textit{Target Branch2} to the \textit{Blending Branch} (\textit{top}), blending the feature representations for each target. \textcolor{gray}{Optional Localization Mask*} (\textit{bottom}) indicates the additional mask that we optionally apply over the cross-attention maps for localization control. The mask identifies local regions described by the selected control vertices and different weights are assigned to each of these regions. For more details, please see Figure \ref{fig:system_local} and Localization Control part of Section \ref{sec:method}.}
\label{fig:bsd}
\end{figure}
%\vspace{3mm}
\medskip
\noindent
\textbf{SDS Guidance for a Single-Target Mesh Deformation.} To stochastically optimize any arbitrary parameters with respect to a pre-trained 2D diffusion model, \cite{poole2022dreamfusion} proposed the Score Distillation Sampling (SDS) process, where given a rendered image $\mathbf{z}$ and a text condition $y$, the objective is to minimize the $L_2$ loss between a sampled noise $\epsilon \sim \mathcal{N}(0, \mathbf{I})$ added to the image, and the noise $\epsilon_\omega$ predicted by a denoising unet $\omega$ at some timestep $t$, sampled from a uniform distribution $t \sim U(0,1)$: 

\begin{figure*}[tb!]
\begin{center}

\centering
    \includegraphics[width=\linewidth]{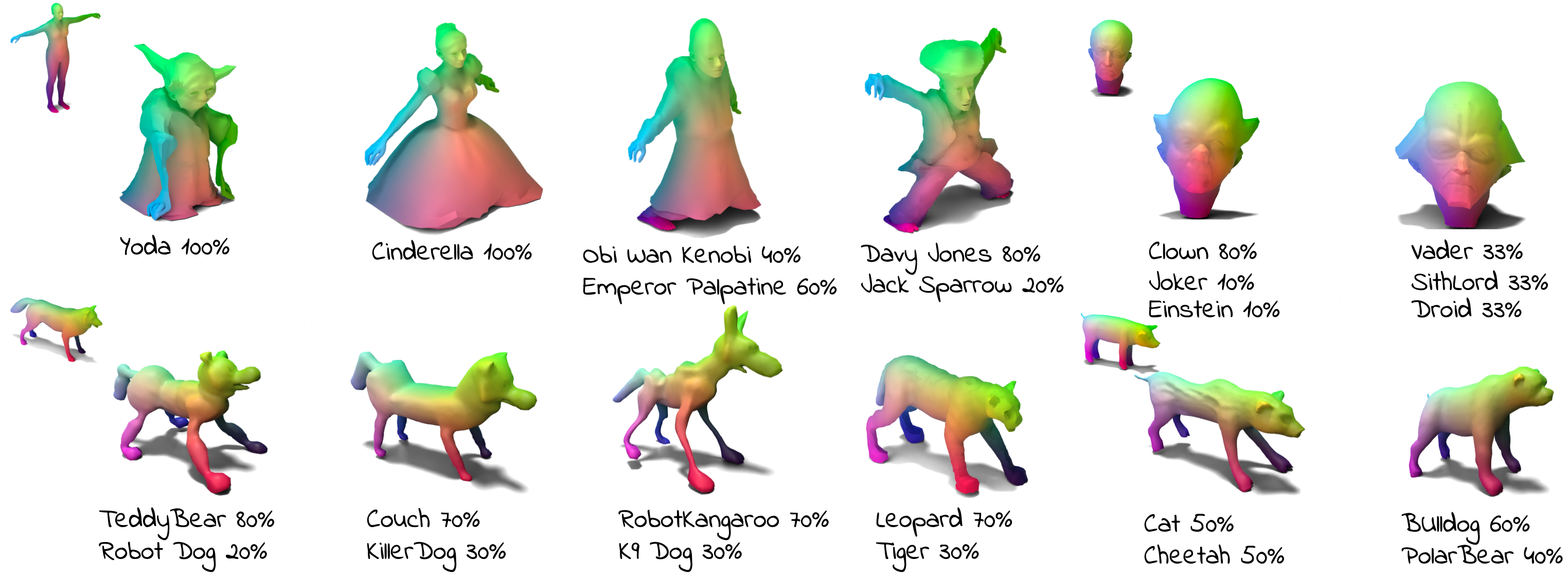}
    
    \vspace{-3mm}
    \captionof{figure}{{\bf Results Gallery.} We present a diverse set of 1-way, 2-way, and 3-way blending results of \ourmethod{}. MeshUp can operate on various kinds of source shapes like human body, face, or animals, and can deform them into a blend of multiple concepts.}
    \label{fig:gallery2}
\end{center}
\end{figure*}
\begin{figure}
\centering
    \begin{overpic}[width=\columnwidth]{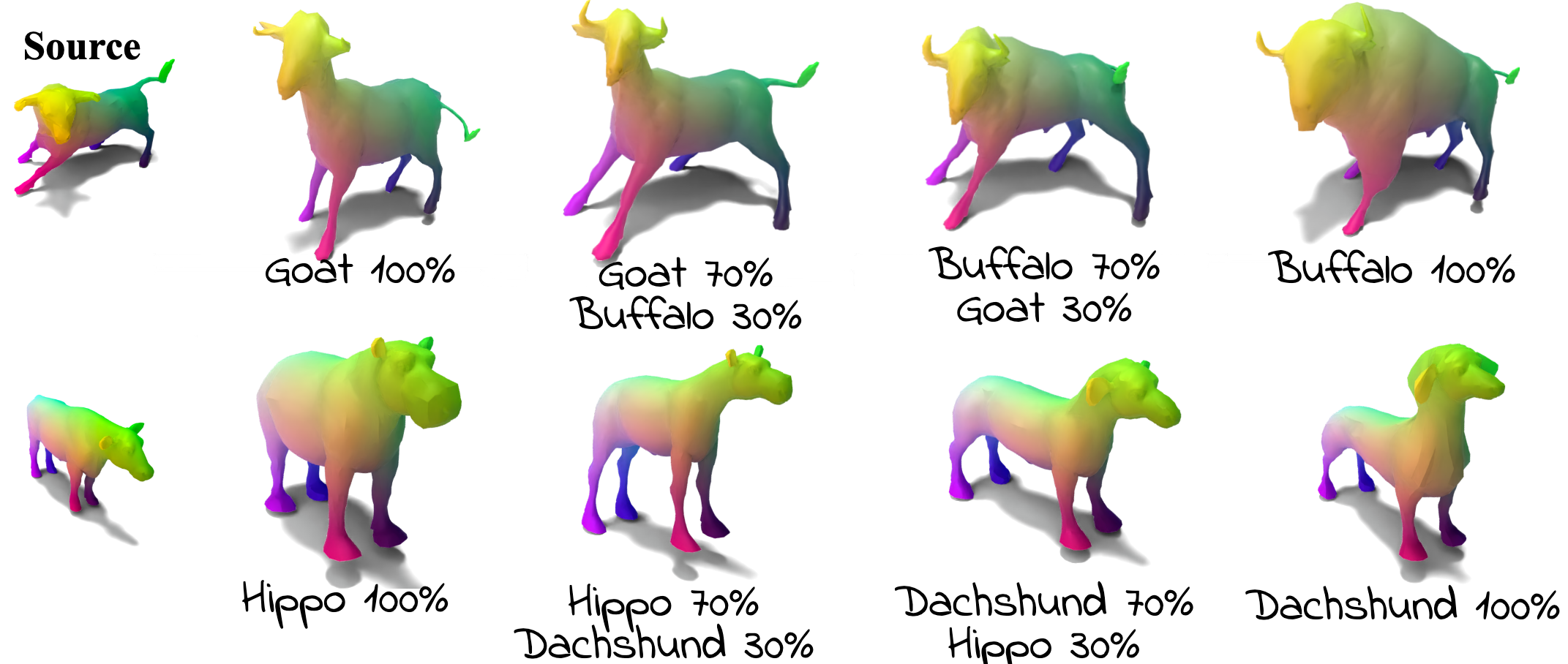}
        % \put (84,18) {\small \textit{Buffalo}}
        % \put (17,18) {\small \textit{Goat}}
        % \put (17,-3) {\small \textit{Hippo}}
        % \put (82,-3) {\small \textit{Dachshund}}
    \end{overpic}
    \vspace{-7mm}
    \caption{{\bf Interpolation Between Two Objectives.} We show that we can vary the ratio between two objectives (\eg going from hippo 100\% on the left to Hippo 70\%-Dachshund 30\%, Hippo 30\%-Dachshund 70\% and finally Dachshund 100\% on the right), effectively interpolating between the shape of the two targets.}
    \label{fig:multipleblend}
    \vspace{-4mm}
\end{figure}

\vspace{-2mm}
\begin{equation}
\mathcal{L}_{\text{Diff}}(\omega, \mathbf{z}, y, \epsilon, t)=w(t)\left\|\epsilon_\omega\left(\mathbf{z}_t, y, t\right)-\epsilon\right\|_2^2,
\label{eq:ldif}
\end{equation}
where $w(t)$ is a weighting term used in the pretrained diffusion model \cite{poole2022dreamfusion}, and \(\mathbf{z}_t\) is the rendered image noised at the timestep $t$. In practice, to compute the gradient of the optimizable parameters efficiently with respect to the loss $\mathcal{L}_{\text {Diff}}$, it has been shown that the gradients through the U-Net of the diffusion model can be omitted \cite{poole2022dreamfusion,sjc}. Since we aim to minimize the loss $\mathcal{L_{\text{Diff}}}$ by optimizing each jacobian $J_i$, we can estimate the gradient of the loss with respect to each jacobian as follows: 
\begin{equation}
\nabla_{J_i} \mathcal{L}_{\mathrm{SDS}}(\omega, \mathbf{z}, y, \epsilon, t) = w(t) \left( \epsilon_{\omega}(\mathbf{z}_t, y, t) - \epsilon \right) \frac{\partial \mathbf{z}_t({J_i})}{\partial J_i},
\label{eq:sds}
\end{equation}

Using this SDS gradient, one can deform a mesh to a single target prompt. A detailed discussion can be found in the supplementary materials.

Following ~\cite{Gao_2023_SIGGRAPH}, we also find it beneficial to regularize the deformation by adding a Jacobian regularization loss 
\begin{equation}
\mathcal{L}_I = \alpha \sum_{i=1} \left\| J_i - I \right\|_2,
\end{equation}

\noindent where $\alpha$ is a hyperparameter determining the regularization strength. This loss penalizes the Jacobians against the identity matrix (which represents the identity deformation) to effectively restrict the magnitude of the deformation. Next, we describe how we extend this framework to multiple targets.

%\vspace{3mm}
\medskip
\noindent
\textbf{Multi-target Guidance via BSD.} Our multi-target architecture is composed of several parallel diffusion branches: one that takes a null text prompt as input (the blending branch), and others with a user-specified target input prompt(the target branches) (see Figure~\ref{fig:system}). These branches also take the same batch of mesh renderings as input images.

For clarity, let j denote the index for the $j^{th}$ target-branch, each associated with a target ``concept." The $j^{th}$ branch would take as input its associated target text, $y_j$, and a weight $w_j$ that controls the degree to which $y_j$ should be expressed. The key observation is that the activation matrices ($\phi^j$) we get at the end of each attention layer represent the``weighted feature space" of each concept, defined over the space of the patch of the input renderings. To blend the features over this space, we perform a weighted interpolation of the activations across the patch dimension, and inject them into the corresponding patch location in the blending branch. Formally, we inject the activation for a single concept as follows:
\begin{equation}
    \phi^{blend} \leftarrow w_j \phi^j + (1-w_j)  \phi^{blend}
\end{equation}

\noindent where $\phi^{blend}$ and $\phi^j$ are the activation matrices in the blending branch and target branch $j$, respectively.
To blend two concepts from the $i^{th}$ and $j^{th}$ target branch, we would inject:
\begin{equation}
    \phi^{blend} \leftarrow w_i \phi^i + w_j \phi^j + (1-w_i-w_j) \phi^{blend}
\end{equation}

The denoising U-Net from the blending branch utilizes the blended activations $\phi^{blend}$ to predict the noise added to the image, and the gradients are backpropagated using Equation (\ref{eq:sds}) to update the Jacobians.

%\vspace{3mm}
\medskip
\noindent\textbf{Localized Control.} 
\label{method:local}
Notably, the BSD pipeline is designed in a way that can incorporate a more fine-grained control over the location where each concept is manifested. Specifically, we select a set of \textit{control vertices} as additional inputs, and impose a novel localization constraint over our concept-mixing pipeline by leveraging the self-attention maps extracted from these vertices. We will first go over how we can achieve local control for a single target, then extend this concept to enable localized blending of multiple concepts.

We first begin by mapping the 3D vertex positions to their corresponding pixel locations in a set of rendered images by using a mapping function $r$ that takes as input $v$, the vertex positions, and $c$, the camera parameters, to find a vertex-to-pixel mapping $P$: 
\begin{equation}
    P = r(v, c).
\end{equation}

\begin{figure}
\begin{center}
    \begin{overpic}[width=\columnwidth]{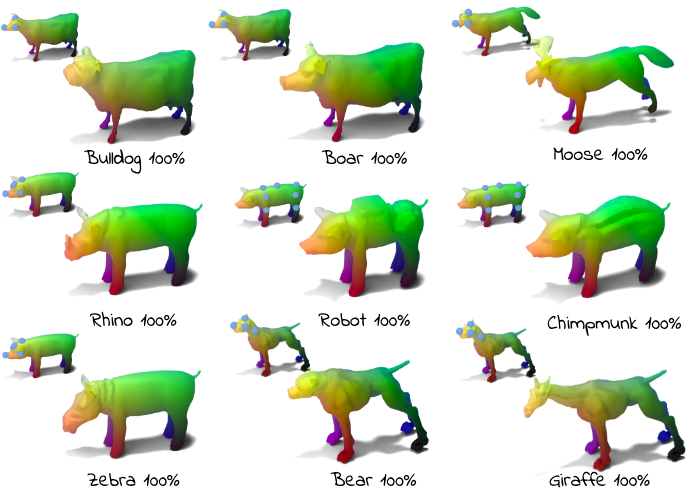}
    \end{overpic}
    \vspace{-6mm}

    \caption{{\bf Local Deformation for a Single Target} 
    We show local deformation results for single text targets. We visualize the source mesh over the top-left corner of each result, and the selected \textit{control vertices} as blue dots. Note how the deformation is constrained to the selected region.}
    \vspace{-7mm}
    \label{fig:singlelocal}
\end{center}
\end{figure}
\begin{figure*}[tb!]
\begin{center}
\begin{overpic}[width=1.\textwidth]{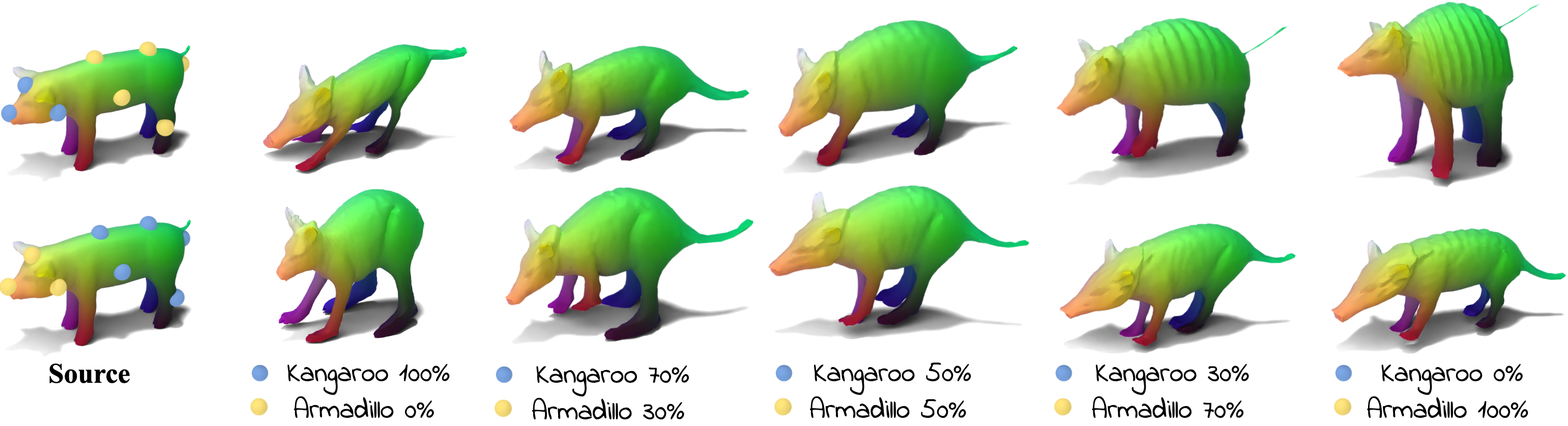}
\end{overpic}
\vspace{-7mm}
\captionof{figure}{{\bf Multiple Local Deformation Control.} We show deformation results for different selection of control points. Both columns show deformation results combined with various BSD weights of 1.0, 0.7, 0.5, and 0.3, 0.0, respectively, for the targets``Kangaroo" and ``Armadillo." \textit{(Top row)} the points around the head region (\textit{blue dots}) are assigned to the target ``Kangaroo," and points around the body to ``Armadillo" (\textit{yellow dots}). \textit{(Bottom row)} flips the assignment. The figure demonstrates how the deformation results vary according to the assignment of selected control points.
}
\label{fig:local_def1}
\end{center}
\end{figure*}

Next, we perform a denoising iteration on these renderings, and using the map $P$, we extract all the self-attention maps corresponding to the selected control vertices. We then average these maps across the attention layers to form a probabilistic region of interest (ROI) for each rendering, which we henceforth denote as $R_m$ (the ROI map for the $m^{th}$ rendering). We then use the inverse pixel map, $P^{-1}$ to map $R_m$ back to its corresponding vertex positions on the mesh surface:
\begin{equation}
    V_R = \sum_m P^{-1}(R_m),
\end{equation}

\noindent where $V_R$ is the 3D probabilistic ROI defined over the mesh vertices. We iteratively update $V_R$ during the BSD optimization process. We can then get a 3D-consistent 2D ROI map, $R_m{'}$, by normalizing $V_R$ with $\hat{V_R} = \frac{V_R - \min{(V_R)}}{\max{(V_R)} - \min{(V_R)}}$, thresholding it at $th=0.8$ to create a binary mask, $mask_{\hat{V_R}}$.

For single-target deformation, we first deform the entire source mesh by regularly updating the jacobians, and at the end of all iterations, we use this binary mask to manually assign any jacobians that falls out of this mask region to the identity matrix:
\begin{equation}
    mask_{\hat{V_R}}(J_i) = I.
\end{equation}
By solving the poison equation ~\ref{eq:poisson} after such assignment, we effectively get a mesh that smoothly deforms to the target only within the region specified by the 3D consistent mask.
As we visualize in Figure~\ref{fig:singlelocal}, our method's significant capability to deform the specified region while preserving its smooth connectivity to the preserved region offers our work to be used as a geometry-editing tool, where given a pre-defined source mesh, the users can select and partially edit specific regions of the mesh using text prompts. 

\medskip
\noindent\textbf{Localized Control for Multiple Concept Blending.} 
\label{method:local}
To ``blend" guidance from a variety of these localized objectives, we first rasterize each of the 3D-mask $mask_{\hat{V_R}}(J_i)$, back to the 2D rendering space, 
\begin{equation}
    R_m{'} = Rast(mask_{\hat{V}_R}, v, c).
\end{equation}

\noindent We then use $R_m{'}$ to mask-out the cross-attention maps, eliminating any association between the target and the unwanted regions of the mesh. Using BSD to mix activation from these masked attention maps yield a guidance score that respects both the weighted blend of multiple targets, as well as their associated local regions, as noticeable in Figure ~\ref{fig:local_def1}

\noindent Since mask $R'_m$ is rasterized from a unified 3D ROI map $V_R$, it is consistent across multiple-viewpoints, and thus for the various renderings. Additionally, because $V_R$ is continuously accumulated as the sum of the attention probabilities projected from multiple $R_m$s, it is guaranteed that the influence from a single attention map is minimized, preventing any particular viewpoints from adding significant variations to the ROI map. We show an ablation of this method in the supplements.

The rasterized $R'_m$ is then used as a binary mask in our usual BSD pipeline to be applied over the cross-attention maps of a desired concept, constraining the area over which the concept can be manifested.
Additionally, since self-attention maps extracted from real, non-inverted renderings can be less informative, we optionally fine-tune and overfit LoRA weights to precisely predict the noise from a large batch of multi-viewpoint renderings using the objective from \cite{ruiz2023dreambooth}. We supply further details about this, as well as the localization method in the supplements.

\medskip
\noindent
\textbf{Image Targets with Textual Inversion.} Text prompts might often be insufficient to describe the desired target and images could be more descriptive in some settings. We leverage textual inversion~\cite{gal2022textual}, which converts an image target into a prompt encoding, and use the encoded prompt in place of the target prompt $y$ of the target branch in our BSD framework.

\section{Experiments}
\label{sec:exp}

\begin{figure}
\begin{center}
    \begin{overpic}[width=\columnwidth]{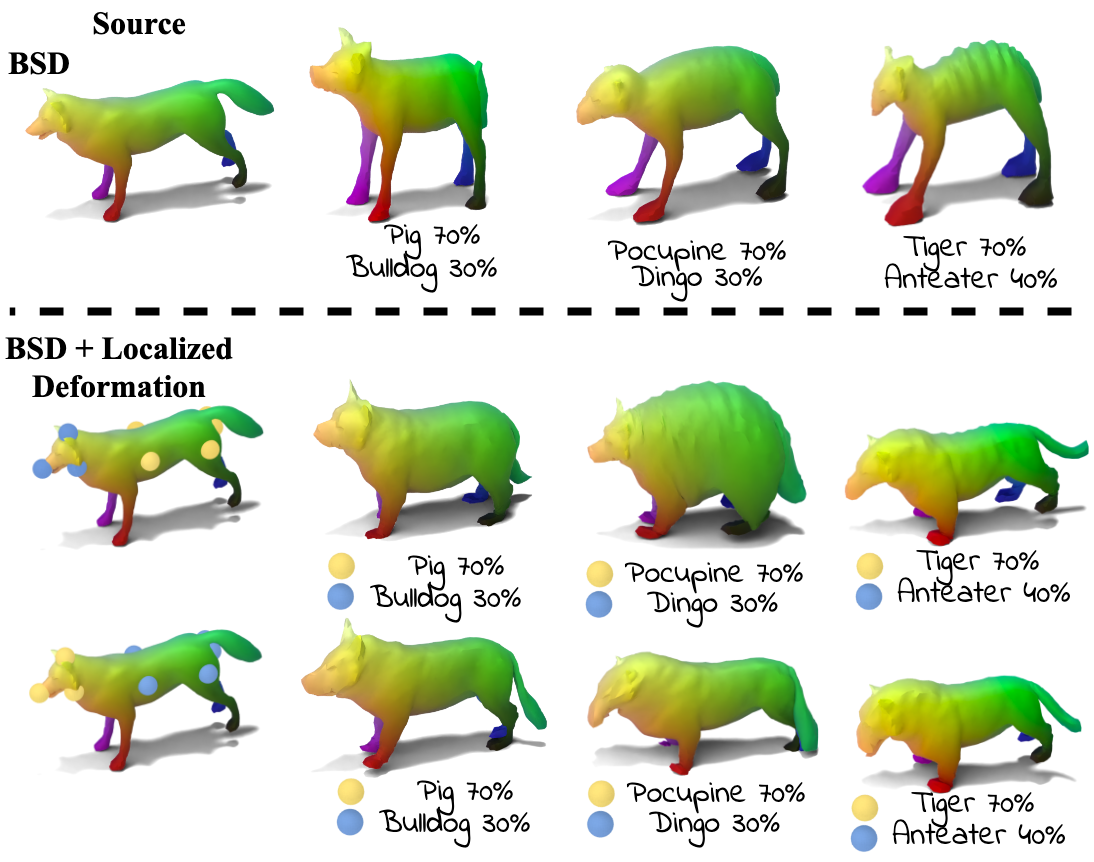}
    \end{overpic}
    \vspace{-6mm}
    \caption{{\bf Evaluation of BSD with and without Local Deformation.} 
    We evaluate the deformation results with and without the Localized Deformation method. The \textit{top row} shows results using just the naive BSD (our regular multi-target deformation), while the \textit{middle} and \textit{bottom row} shows the results using our localization method. We visualize the selected \textit{control vertices} as blue and yellow dots on the mesh. Note how the results using our Localized Deformation method respect the assigned \textit{control points}, in addition to the mixture of multiple targets.}
    \vspace{-6mm}
    \label{fig:local}
\end{center}
\end{figure}
\begin{figure*}[tb!]
    \centering
    \begin{overpic}[width=\linewidth]{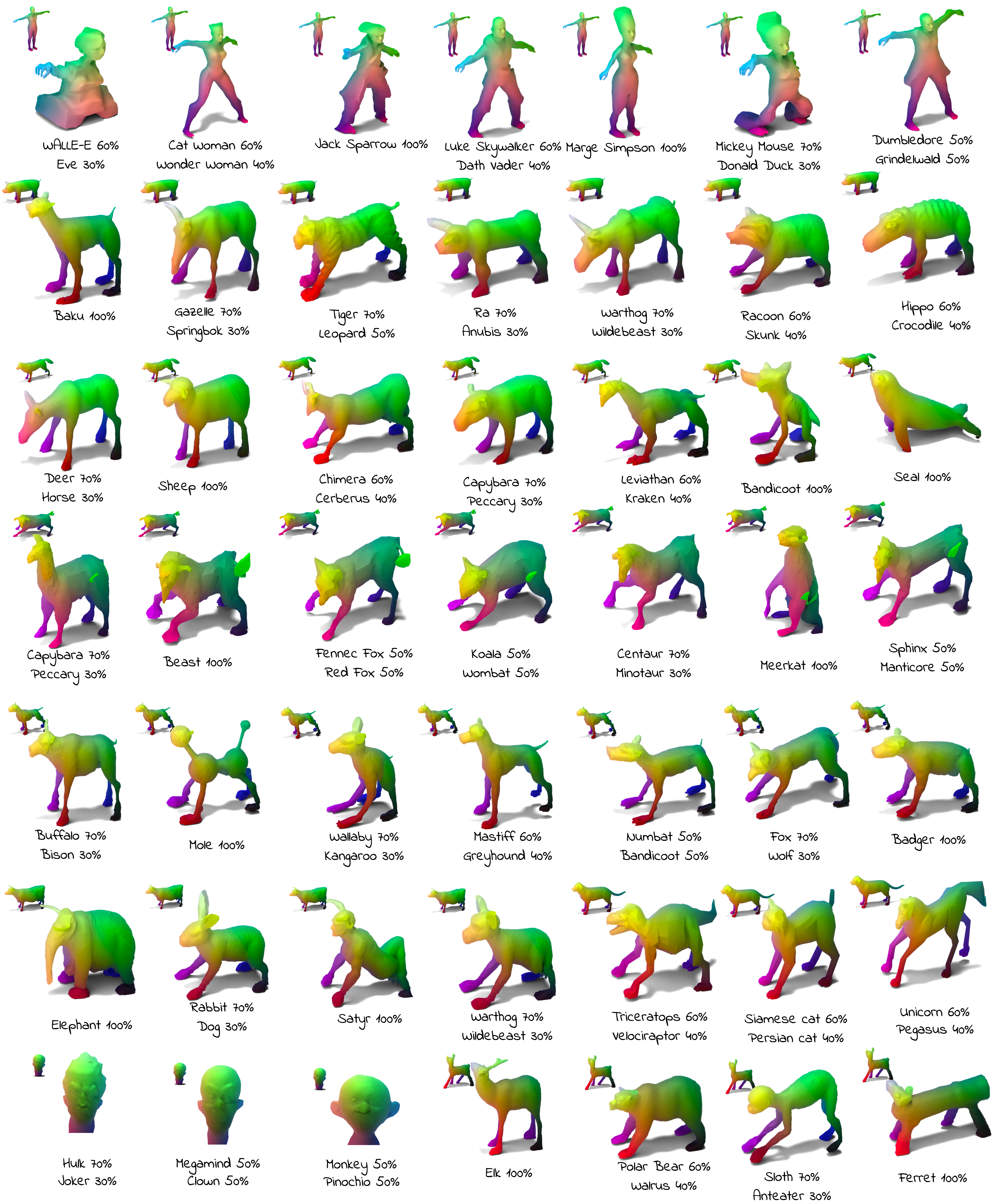}
    \end{overpic}
    \vspace{-3mm}
    \caption{\textbf{Gallery.} We present a diverse set of 2-way \ourmethod{} deformation results.}
    \label{fig:supp_gallery}
\end{figure*}

In this section, we first show multi-target deformation results driven by text or image targets. Additionally, we demonstrate deformations with local control and mesh targets. Finally, we also describe how our method can be used as a regularization term that controls the strength of a deformation. We provide comparisons with various baselines, show an experiment that uses our method to perform keypoint interpolation between concepts, and show a qualitative user study of our method in the supplementary material. We also provide a comprehensive collection of various deformation results in Figure~\ref{fig:supp_gallery}.

\subsection{Concept Mixing Results}

\begin{figure}[tb!]
\begin{center}
\centering
    \includegraphics[width=\linewidth]{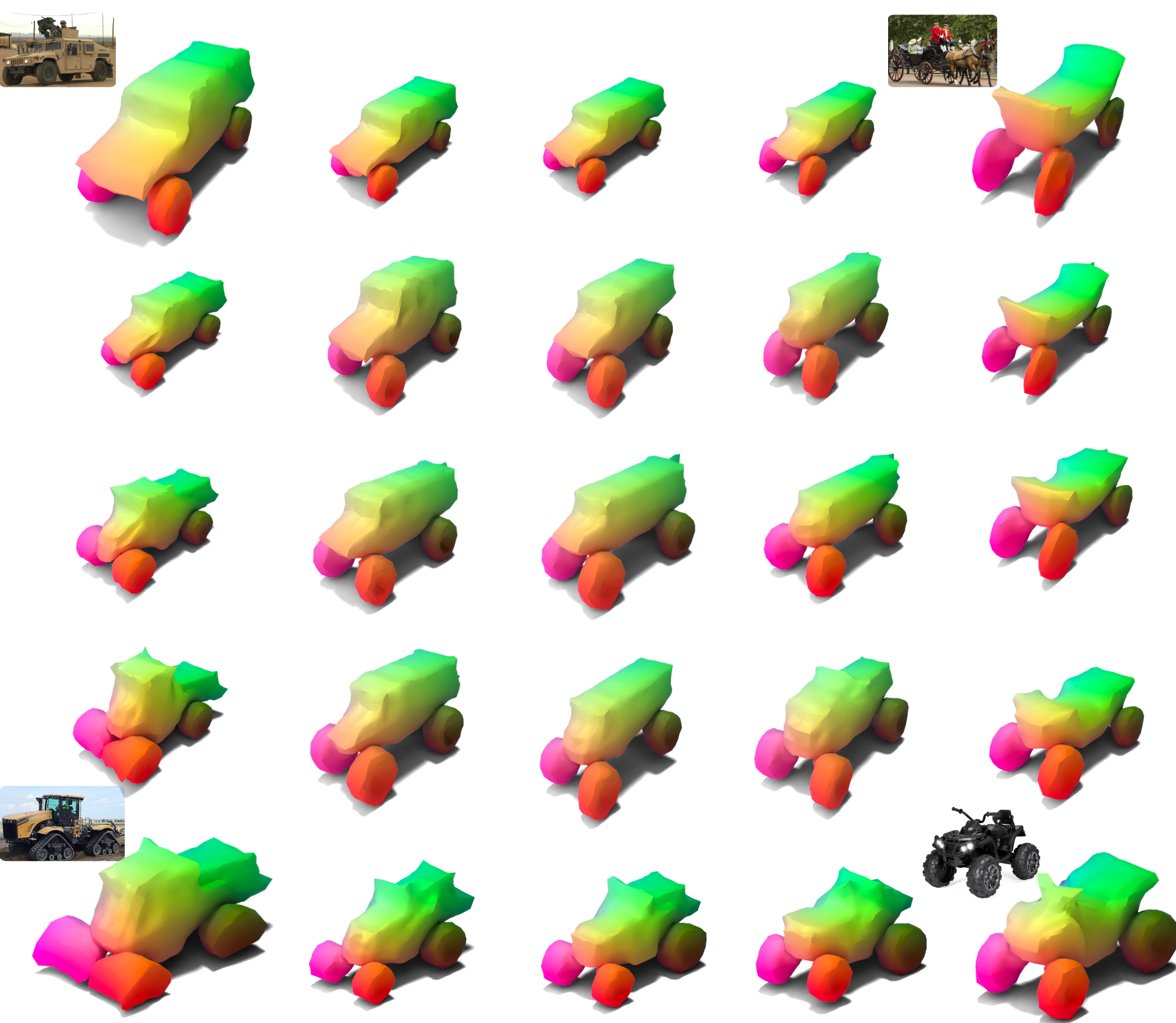}
    \captionof{figure}{{\bf Four-way Blending with Image Targets.} We use textual inversion~\cite{gal2022textual} to condition the Text-to-image model with inspiration images represented by their inverted textual token. \ourmethod{} supports as many targets as desired. We demonstrate a four-way blending with four target image concepts. The closer the shape is to the target image, the higher the corresponding blending weight of that target.}
    \label{fig:gallery}
    \vspace{-7mm}
\end{center}
\end{figure}

\medskip
\noindent \textbf{Multi-Target Results.}
We demonstrate various multi-target concept mixing results in Figures~\ref{fig:teaser}, ~\ref{fig:gallery2}, and ~\ref{fig:multipleblend}. Our method successfully mixes diverse concepts (animals, faces, fantasy creatures, and vehicles) with various weights. The figures illustrate how the same concepts can be mixed with different weights, enabling the user to control which features emerge more prominently. For example, in Figure~\ref{fig:multipleblend}, we can clearly see how with a high weight for the hippo shape, the fat body and the rounded face is prominent. On the other hand, dachshund's long body and facial features are dominant for examples with higher weights on dachshund.

\begin{figure}
\begin{center}
\centering
    \begin{overpic}[width=\columnwidth]{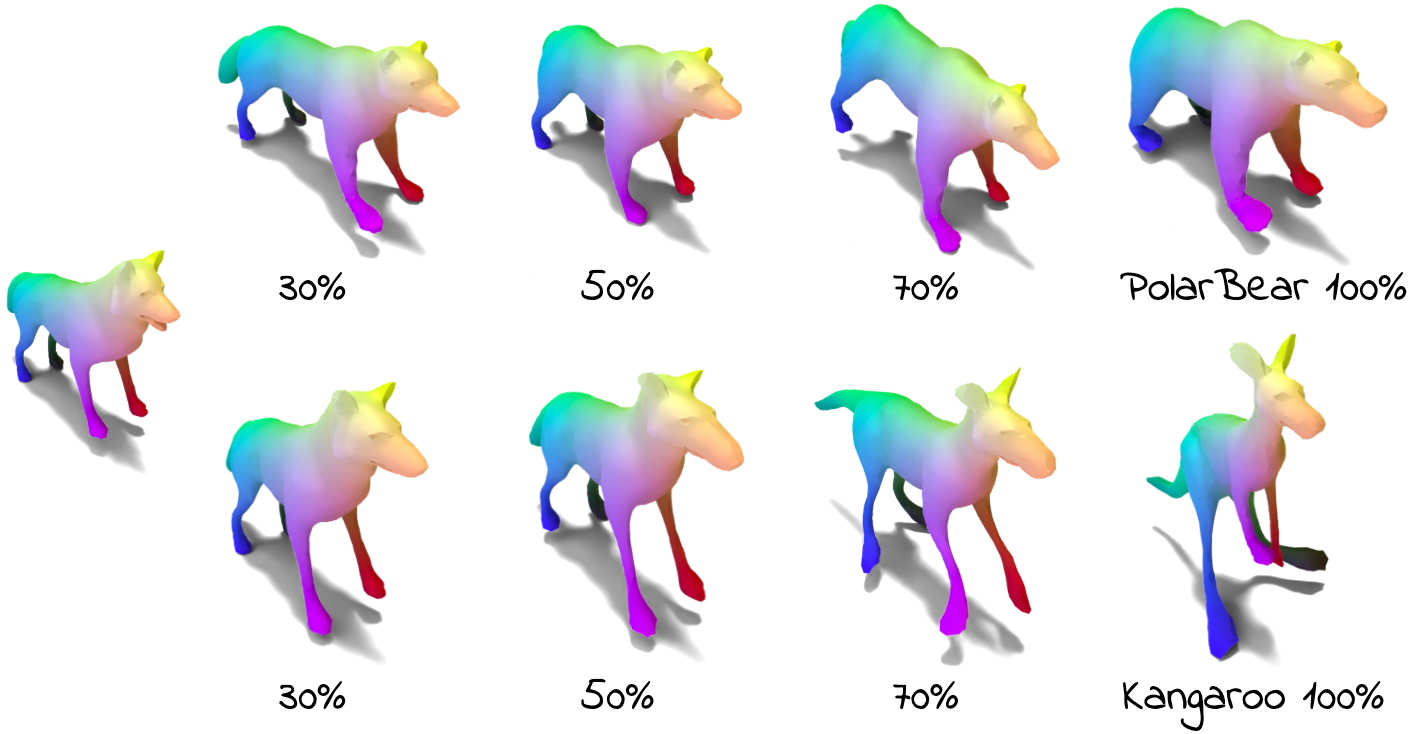}
        
    \end{overpic}
    \vspace{-8mm}
    \caption{{\bf Self-Blending Deformations}. We use BSD to inject varied strengths of activations, each from a single target to the blending branch, effectively controlling for the degrees to which the target is expressed in the resulting deformation.}
    \vspace{-3mm}
    \label{fig:self_blend}
\end{center}
\end{figure}

\medskip
\noindent \textbf{Localization Control Results.}
In Figure~\ref{fig:singlelocal} and Figure~\ref{fig:local_def1}, we provide examples of localization control, where users can indicate (by selecting the control vertices, visualized in blue and yellow dots) which part of the model should be affected by each target. Note how each of the target features emerges in the user-specified region. This method offers a high level of control over how both mixed/unmixed concepts manifest in the deformed mesh. We also demonstrate how the local deformation is affected by changes in the assignment of weight ($w$) in Figure~\ref{fig:local_def1}. We observe that if we give a different emphasis to different parts of
the source mesh via the selection of control vertice,
BSD conditions the emphasis accordingly on various scales, creating a
more versatile space for user control.

\medskip
\noindent \textbf{Image Targets and the Concept Space.}
We show the ability of our system to take image concepts as inputs in Figure~\ref{fig:gallery}. We find this feature to be especially useful for some concepts that have significant shape variations (e.g., ``trucks''), and for those that are difficult to engineer a precise prompt for. In this figure, we also illustrate how one can generate a continuous blending space spanning as many as four concepts by sampling different relative weights for each one of them.

\medskip
\noindent \textbf{Regularizing Mesh Deformation via Self-Blending.}
\begin{figure}
\begin{center}
    \vspace{-2mm}
    \begin{overpic}[width=\columnwidth]{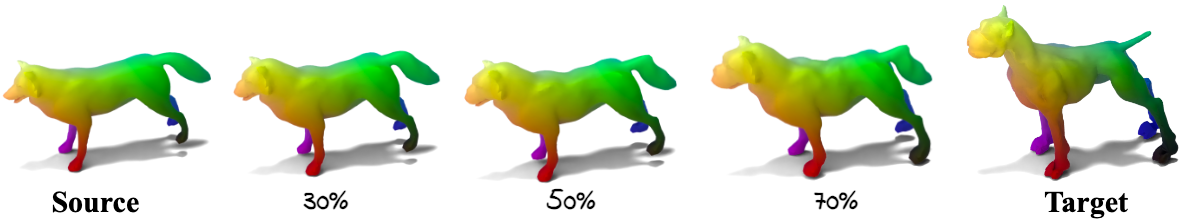}
    \end{overpic}
    \vspace{-5mm}
    \captionof{figure}{{\bf Interpolating Mesh Using Self-Blending Deformation.} 
    We show how the Self-Blending capability of MeshUp can be used to interpolate the shapes of two meshes, the \textbf{Source} and the \textbf{Target}, by using dreambooth to learn the shape of the \textbf{Target}, and deforming the \textbf{Source} using various weights. Note how the muscular features of the \textbf{Target} mesh gradually emerge as we increase the blending weight from 30\% to 70\%.}
    \vspace{-5mm}
    \label{fig:meshinterp}
\end{center}
\end{figure}
\begin{figure}
\begin{center}
    \vspace{-3mm}
    \begin{overpic}[width=\columnwidth]{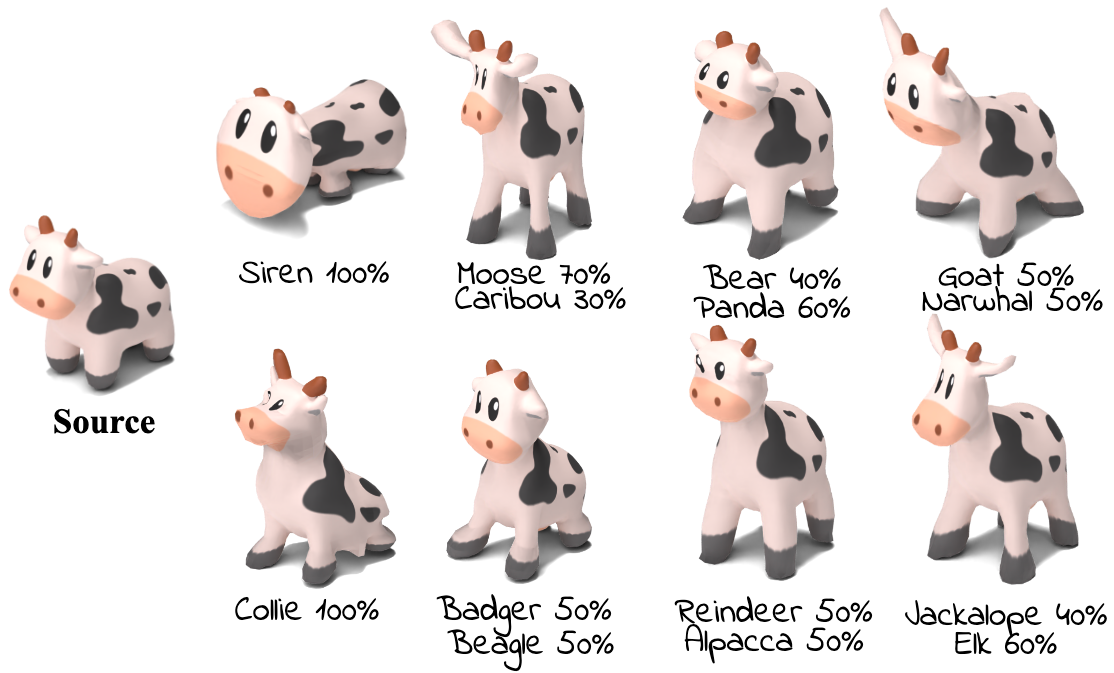}
        
    \end{overpic}
    \vspace{-8.5mm}
    \captionof{figure}{{\bf Texture Transfer.} 
    We show how the texture map initially defined over the source mesh gets transferred without distortion to meshes deformed using our method.}
    \vspace{-9mm}
    \label{fig:texture_transfer}
\end{center}
\end{figure}
In Figure~\ref{fig:self_blend} we demonstrate that our BSD pipeline can take a single target objective, and be used to control the strength of a single-target deformation by using various weights, $w_j$. We use a modified classifier free guidance to achieve this (we provide its details and ablation in the supplements). 
\begin{figure*}[h]
\begin{center}
    \vspace{-5mm}
    \begin{overpic}[width=\textwidth]{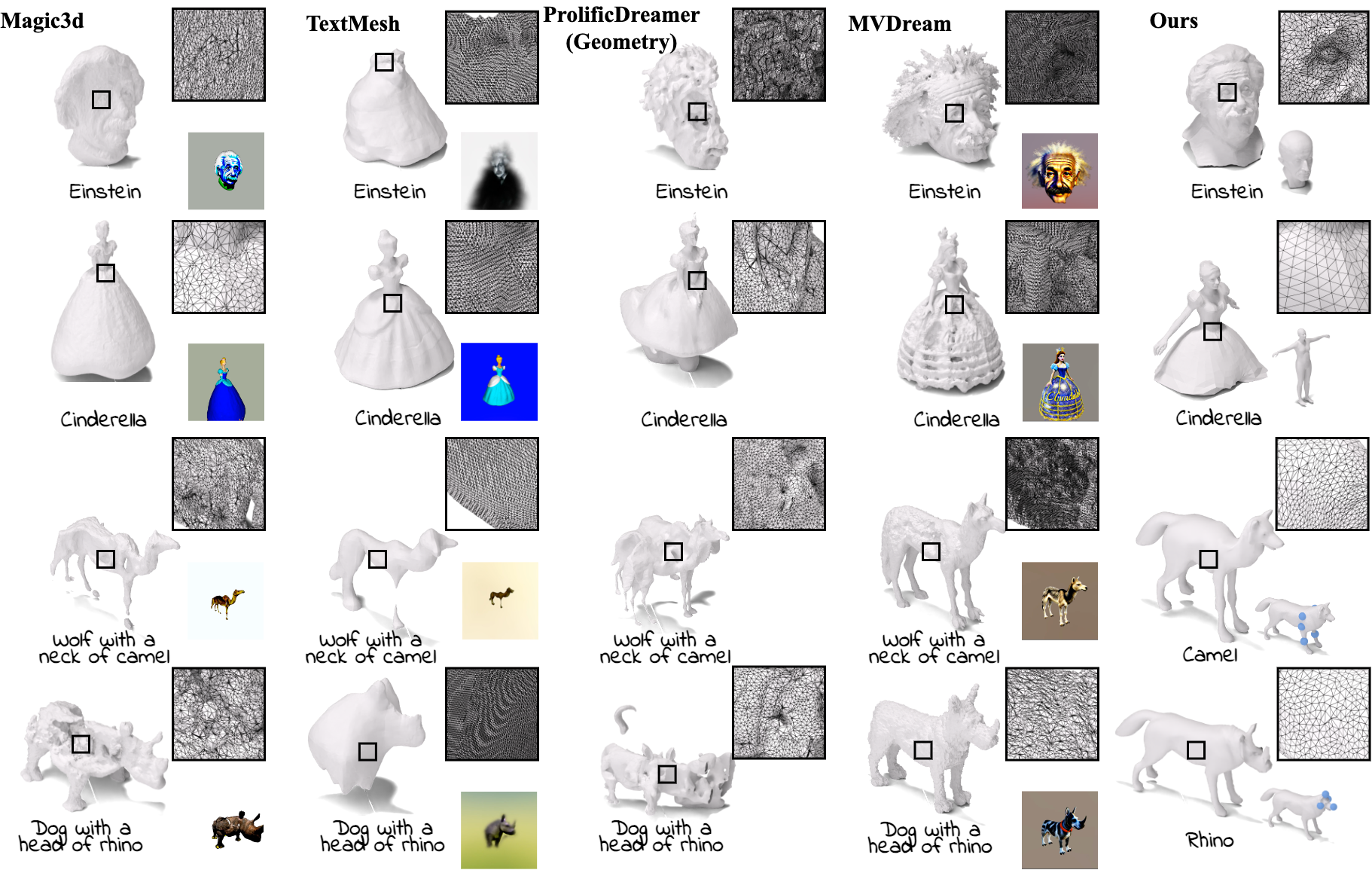}
        
    \end{overpic}
    \captionof{figure}{{\bf Comparison with other methods.} 
    We compare the mesh quality obtained with \ourmethod{} to Magic3D, TextMesh, ProlificDreamer, and MVDream. For Magic3D, TextMesh, and MVDream, we visualize the textured, implicit shape representation on the bottom right of each figure (for ProlificDreamer, we only show the result after the geometry refinement stage). For our first two results, we deform the source mesh (visualized on the bottom right of each result) into the specified targets. For the last two results, we use our localized control method to confine the deformation to the region specified by the control vertices (visualized as blue dots over the source mesh).}
    \label{fig:dreamfusion_comp}
    \vspace{-9mm}
\end{center}
\end{figure*}

\medskip
\noindent \textbf{Using Mesh as Targets.}
In Figure~\ref{fig:meshinterp}, we further expand our model's self-blending capability to interpolate the shapes of two meshes, by gradually deforming the source mesh into a \textbf{Target} mesh. To achieve this, we utilize the multi-viewpoint renderings of the target, and batch 48 renderings per-iteration to fine-tune the UNet of the diffusion model using the objective from DreamBooth~\cite{ruiz2023dreambooth}. To avoid memory overload, we fine-tune the LoRA weights~\cite{hu2021lora} instead of the whole model. Using the fine-tuned weights with the associated token as the objective, we deform the \textbf{Source} using various weights, $w_j$. Please refer to ~\cite{ruiz2023dreambooth} for details of the training procedure.

\medskip
\noindent \textbf{Texture Transfer.}
We demonstrate the utility of deforming from a source shape using our method, as opposed to generating new 3D shapes from scratch. In Figure~\ref{fig:texture_transfer} we show how the texture map defined over the source seamlessly transfers over to other meshes deformed using our method. We can extend this property to transfer other attributes such as motion functions, and we show this example in the supplementary video, which can be found in our project page.

\medskip
\noindent \textbf{Comparison with Other Methods.}
Finally, we compare the quality of our mesh outputs to those extracted from Magic3D, TextMesh, ProlificDreamer (geometry refinement stage), MVDream. Not only does our method yield a geometry of much better detail and quality, but the tessellations (visualized on the right side of each figure) are also superior, a crucial advantage for any mesh-based graphic applications. We also show in the last two \textit{bottom rows} that our localized control method significantly outperforms other methods that use text description to depict the localized deformation results we can achieve using MeshUp. More details of the comparison, including the specific models we used for these experiments, can be found in the supplements.

\vspace{-1mm}
\section{Conclusion and Limitations}
In this paper, we propose a versatile framework for mesh deformation that supports deformation via text or image-based concepts, mixing these concepts using various weights, and localizing their expressions.

The deformation of \ourmethod{} is focused on preserving the topology of the initial mesh, and treating it as a shape-prior which prescribes the aesthetic of the deformed results. Thus, our technique would not be suitable for inducing topological changes (such as deforming a sphere to an object with topological holes). We leave the task of topology-modifying mesh modifications to future work. Although we limit our focus to deformation in this paper, another potential application of our method would be to leverage our technique for generating other mesh parameters, such as textures, materials, and normals.

{
    \small
    \bibliographystyle{ieeenat_fullname}
    \bibliography{bibs}
}
\newpage
\clearpage
\appendix

\section*{Supplementary Material}
The following sections will provide more details of our method, present additional experiments, and evaluate our method using quantitive/qualitative metrics. Section A will delineate the technical and mathematical details of our method, including model configurations, runtime, Score Distillation Sampling (SDS), localized deformation, and the modified classifier free guidance score used in our self-blending experiments. Section B will present the experiments, and Section C will provide evaluations. 

\section{Technical Details}

First, we briefly justify our choice of DeepFloyd-IF XL model as our denoiser. For our purposes, we noticed that IF performs significantly better than the widely used latent Diffusion models, most likely as it operates directly in the rgb space. In addition to the denoiser, the t5 text encoder on which IF is trained has a much more expressive embedding space than that of CLIP. Since we aim to employ Textual Inversion to learn a set of images through this embedding space, the t5 encoder comes to our great advantage. We leave more extensive comparison across various diffusion models to future work.

For runtime, we ran all of our results using a single A40 gpu, and ran 2400 iterations with a batch size (number of renderings per iteration) of 16. It takes approximately 1.5 hours to 
optimize for a single target. The time increases by a factor of approximately 1.5 
for each target that we add for blending. However, we have observed that the deformation converges reasonably close to the final results after 600-800 iterations, which takes about 25-30 minutes to run for a single target. Running the localized deformation method also increases the run-time approximately by an factor of 1.5, not including the optional LoRA-finetuning stage, which takes an extra 10-15 minutes to converge. The optimization time also depends heavily on the number of faces of the source mesh, and we can therefore only provide a rough approximation of the run-time. Most of the meshes that we used in this paper had 10,000-30,000 faces.

For rendering, we use nvdiffrast to rasterize the mesh from multiple viewpoints. Since our main objective is to optimize the geometry of the source mesh (and not its texture), we simply paint the mesh with a uniformly grey (0.5, 0.5, 0.5) texture, and rasterize it to get the renderings. We found the grey texture works reasonably well for our purpose, but we leave the investigation of more sophisticated texturing schemes up for future work. We also rasterize the mesh in 512x512 resolution and downsample it to 64x64 using bilinear interpolation before inputing it to the diffusion model. 

\vspace{3mm}
\noindent
\textbf{Specific Overview of SDS} We will go over the SDS perspective of diffusion in detail, and how we apply this perspective for our purpose of deforming the Jacobians.
We first limit our scope to 2D image-to-image generation, where, given an input image $\mathbf{z}$, the objective is to optimize some parameter $\theta$ in the 2D space.
Given a text condition $\mathbf{t}$, we define the loss function to be the squared L2 norm between the noise predicted by the denoising model $\epsilon_\phi$ and the sampled noise  $\epsilon \sim \mathcal{N}(0, \mathbf{I})$:

\begin{equation}
\mathcal{L}_{\text{Diff}}(\phi, \mathbf{z}, y, \epsilon, t)=w(t)\left\|\epsilon_\phi\left(\mathbf{z}_{\mathbf{t}}, y, t\right)-\epsilon\right\|_2^2,
\end{equation}

where $t$ is the randomly sampled timestep from a uniform distribution, $t \sim \mathcal{U}(0,1)$, $\epsilon$ is a random noise sampled from a standard normal distribution, and $\mathbf{z}_{\mathbf{t}}$ is the input image noised according to timestep $t$ using the re-parameterization trick, $\mathbf{z}_{\mathbf{t}}=\sqrt{\alpha_t} \mathbf{z}+\sqrt{1-\alpha_t} \epsilon$. $w(t)$ is a weighting function for which we will not go into details. In simple terms, we aim to optimize some parameters so that the (frozen) model can precisely predict the noise sampled from timestep $t$. When this loss is minimized, the parameters ($\theta$) are optimized to represent an object that is as close as possible to what is predicted by the denoiser to be of highest probablity. A deeper analysis of diffusion models and Score Distillation Sampling can be found in \cite{DDPM, scorejacobian, poole2022dreamfusion}

The gradient of $\mathcal{L}_{\text{Diff}}$ with respect to $\theta$, which we denote $\nabla_\theta \mathcal{L}_{\text{Diff}}$, can be derived by,

\begin{equation}
\nabla_\theta \mathcal{L}_{\text{Diff}}=\left(\epsilon_\phi\left(\mathbf{z}_{\mathbf{t}}, y, t\right)-\epsilon\right) \frac{\partial \epsilon_\phi(\mathbf{z}, y, t)}{\partial \mathbf{z}_{\mathbf{t}}} \frac{\partial \mathbf{z}_{\mathbf{t}}(\theta)}{\partial \theta} .
\end{equation}

It is known from \cite{poole2022dreamfusion} that instead of having to backpropagate through the denoiser we can approximate an effective gradient for $\theta$, denoted as $\nabla_\theta \mathcal{L}_{\mathrm{SDS}}$, simply by omitting the gradient with respect to the denoiser, $\frac{\partial \epsilon_\phi(\mathbf{z}, y, t)}{\partial \mathbf{z}_{\mathbf{t}}}$, giving us,

\begin{equation}
\nabla_\theta \mathcal{L}_{\mathrm{SDS}}(\mathbf{z}, y, \epsilon, t)=(\epsilon_\phi\left(\mathbf{z}_{\mathbf{t}}, y, t\right)-\epsilon) \frac{\partial \mathbf{z}_{\mathbf{t}}(\theta)}{\partial \theta}
\label{eq:SDS_equation}
\end{equation}

In addition to SDS, we also use classifier-free guidance with an extremely high classifier-free guidance weight of 100 to help with our 3D objective \cite{poole2022dreamfusion}. 

\textbf{Using SDS to guide Mesh Deformation}
In order to optimize for the Jacobians $\mathit{J_i}$ of mesh faces instead of some arbitrary parameter $\theta$, we can simply replace $\theta$ in (\ref{eq:SDS_equation}) with $\mathit{J_i}$,

$$
\nabla_\mathbf{J_i} \mathcal{L}_{\mathrm{SDS}} = (\epsilon_\phi\left(\mathbf{z}_{\mathbf{t}}, y, t\right)-\epsilon)\frac{\partial \mathbf{z}_\mathbf{t}(\mathbf{J_i})}{\partial \mathbf{J_i}}. 
$$

We denote $\mathbf{z}_\mathbf{t}$ to be dependent on $\theta$ since we denote it to encompass the span of the entire computational pipeline, from the optimized 3x3 Jacobians to the renderings of the deformed mesh (this includes the operations of projecting the 3x3 Jacobians to the 3x2 space, running the poison solve to calculate the deformation map $\phi$, deforming the mesh, rendering this deformed mesh onto the image space).
In practice, we use Pytorch's autograd library to automatically handle the differentiation of $\mathbf{z}_\mathbf{t}(\mathbf{J_i})$.

\noindent \textbf{Localized Deformation}
We will now detail the implementation of our localized deformation experiment. This explanation is specific to the DeepFloyd IF model that we use

First, we note that the UNet denoiser consists of multiple  attention layers, and each of these layer takes as input the text encoding and the hidden states, output by the previous layer. We then project the hidden states each as Key, Query, and Value matrices using learned mlp layers. These matrices are then concatenated with matrices similarly projected from the text enbeddings, giving us the matrices,  $Q$, $K$, $V$. \cite{chen2021crossvit, NIPS2017_3f5ee243}
Specifically, the attention map is defined as

$$
M = \text{Softmax}\left( \frac{QK^T}{\sqrt{d}} \right)
$$
While the activation is 
$$
\phi = MV.
$$

The attention matrix $M$ consists of both the self- and cross-attention map. The cross-attention map encapsulates the correlation between the text embeddings and the various ``patches" of
the image, while the self-attention map finds the correlation across these patches.
Using this observation, we first utilize the self-attention maps to identify the regions on the image that have high correspondence with the \textit{control vertices}, and convert this map into a 2D Region of Interest (ROI) mask, $R_m$. 
We then extract the 3D-consistent ROI mask $R'_m$ using our rasterization function, project it back onto 2D, and use it to mask out the cross-attention map of each target. Effectively, we are distilling the regions within the image where target concepts should be expressed. 

We will now delineate some of the implementation details of the localized deformation method. First, as briefly mentioned in the method section, we aim to replicate the effect of inverting the renderings by finetune the denoising UNet of our DeepFloyd IF model by the objective proposed in ~\cite{ruiz2023dreambooth}. While dreambooth finetunes the denoiser so that it can reproduce the input image, we do so for multi-viewpoint mesh renderings, encouraging the model to be capable of reconstructing the mesh renderings given a predefined token. Moreover, since fine-tuning the entire UNet as implemented in the original Dreambooth paper is prohibitively expensive, we instead finetune the LoRA weights ~\cite{hu2021lora} with the same training objective as described above. We run 150 fine-tuning iterations on a batch of 16 images, and the entire operation takes approximately 15 mins. 
While this finetuning process betters the quality of the localization map, we also observe that the map is reasonable enough even without finetuning. Users who desire faster generation time over better localization could skip this process with minimal compromise in quality. 

Additionally, we use the finetuned weights to stabilize the localized deformation method. Notably, when a particular weight $w_i$ is extremely low (over 0.2) for a certain target, the blended activation $\phi^{blend} = w_i\phi^{i} + w_j\phi^{j}$ (where $i$ and $j$ each represent different target concepts) might become extremely unstable as the localized region corresponding to the $i^{th}$ concept could only receive minimal activation signals. In order to compensate for this instability, whenever the $(1-{max}(w_j, w_i))<0.2$ (when one target has extremely small weight), we create an inverse boolean mask, $\sim R'_m$ and apply this mask to the cross attention map of the finetuned UNet. We then weight this masked cross attention by $w_{lora}=0.8$, and simply add it to the cross-attention mask of the $i^th$ target. This operation ensures that all the regions of the mesh receive at least a minimal amount of attention required to keep the activation $\phi^{j}$ stable throughout the deformation process. 

With the localization process, the entire deformation process takes about 2 hours to complete (for a 2-way blending of concepts).

\begin{figure*}[tb!]
\begin{center}
    \begin{overpic}[width=\textwidth]{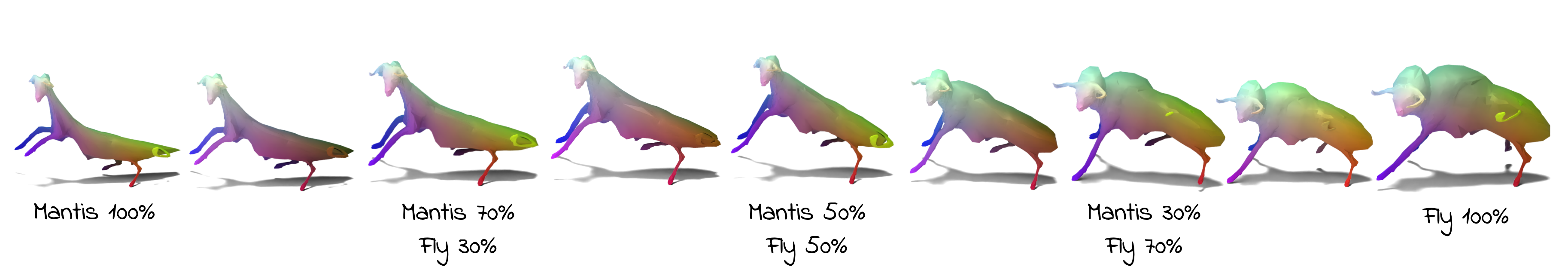}
    \end{overpic}
    \captionof{figure}{{\bf Keyframe interpolation.} We create a continuous combinatorial space of blends by running our method for a discrete number of keyframes and interpolating their vertices to obtain the intermediate shapes in between (the ones without text below). Our correspondence-preserving deformation enables a smooth transition between the keyframes.}
    \label{fig:keypoint}
\end{center}
\end{figure*}

\begin{figure*}[tb!]
\begin{center}
    \vspace{-1mm}
    \begin{overpic}[width=\textwidth]{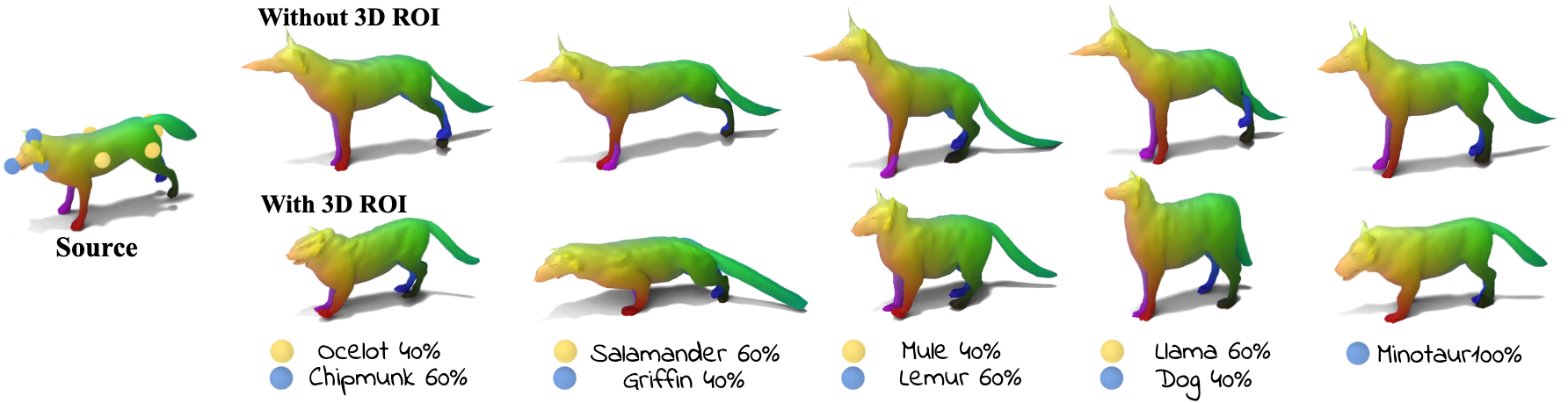}
        
    \end{overpic}
    \vspace{-8mm}
    \captionof{figure}{{\bf 3D ROI map Ablation.} We show an ablation of the 3D ROI map $V_R$ for our localized deformation method. (\textit{top row}) shows the results of using just the 2D ROI maps, $R_m$, extracted from the self-attention maps of each viewpoint, as masks for the cross-attention map. (\textit{bottom row}) uses $R'_m$, the 3D-consistent masks extracted from the 3D ROI map $V_R$.}
    \label{fig:localdef_ablation}
\end{center}
\end{figure*}

\medskip
\noindent \textbf{Modified Classifier Free Guidance for Self-Blending} In this section, we will discuss how we modified the Classifier Free Guidance (CFG) for our self-blending experiments (Figure~\ref{fig:self_blend}).
In the self-blending experiment, we utilize our blending pipeline to control the expression of a single concept. We thus reduce the branches to two (one target branch and one blending branch), and extract the activation as $\phi ^{blend} = w\phi + (1-w)\phi^{null}$, where $\phi^{null}$ is the activation from the blending branch (activation from the null text promt, ``'').

One caveat to this method, however, is that $\phi^{null}$ could introduce undesired bias for $\nabla_{theta}L_{SDS}$, especially when its corresponding weight, $1-w$, is high. 

To tackle this problem, we slightly modify our equation for CFG The original CFG, a method introduced by \cite{ho2022classifierfree}, is formulated as follows:

$$
\begin{aligned}
\epsilon_\phi(\mathbf{z}_{\mathbf{t}}, y, t)=\hat{\epsilon}_{text} + \alpha (\hat{\epsilon}_{text} - \hat{\epsilon}_{null}),
\end{aligned}
$$
where $\hat{\epsilon}_{text}$ and $\hat{\epsilon}_{null}$ are the predicted noise conditioned on the target text prompt and null text prompt, respectively. $\alpha$ is the Guidance Scale, a parameter that controls the strength of CFG. We modify our equation into
$$
\begin{aligned}
\epsilon_\phi(\mathbf{z}_{\mathbf{t}}, y, t)=\epsilon + \alpha (\hat{\epsilon}_{text} - \hat{\epsilon}_{null}),
\end{aligned}
$$
where we simply replace $\hat{\epsilon}_{text}$ with $\epsilon$, the sampled noise.

\begin{figure*}[tb!]
\begin{center}
    \begin{overpic}[width=\textwidth]{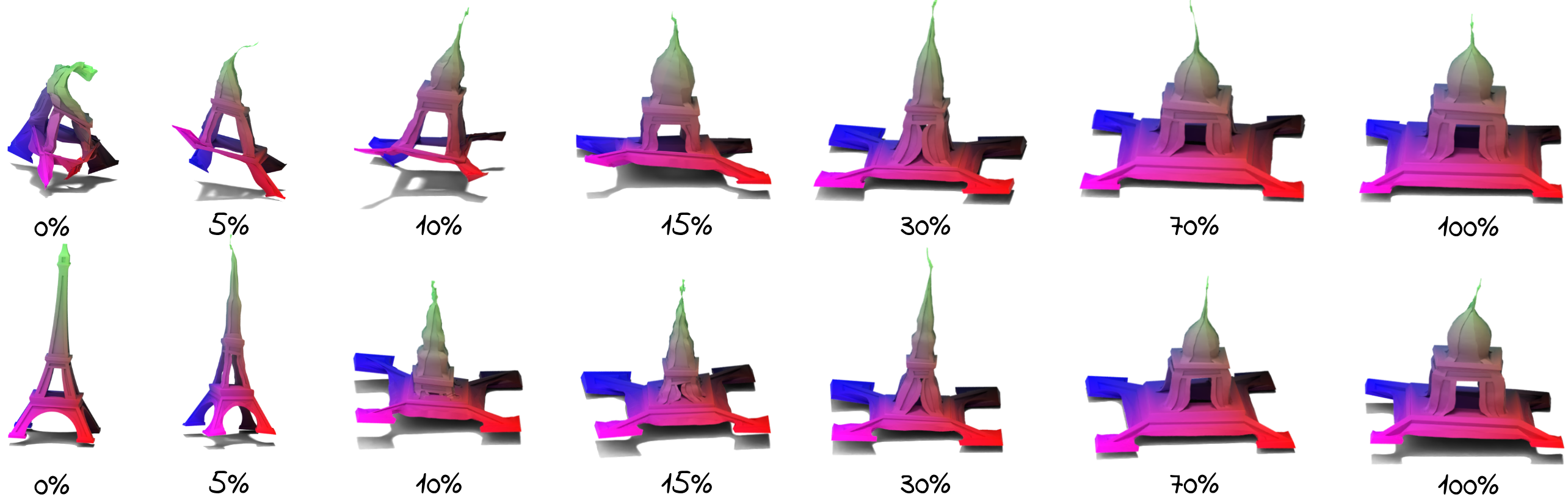}
    \end{overpic}
    \captionof{figure}{{\bf Modifying the classifier-free guidance ablation.} We show an ablation for the modified classifier-free guidance version of our method for single-target self-blending deformation. Results are shown for various blending scales, ranging from 0\% to 100\%. The top row is the regular classifier-free guidance and the bottom one is with our modified version. The regular classifier-free guidance creates artifacts and does not reflect well the mixing percentage. In contrast, our self-blending scheme yields a smooth transition from the source shape to the target deformation objective.}
    \label{fig:mod_cfg}
\end{center}
\end{figure*}

% - what this allows us to do 
Such modification allows our model to achieve a more stable, unbiased control of the deformation strength, as opposed to the original CFG, as shown in Figure~\ref{fig:mod_cfg}. 
Intuitively, we want the result to identity (no deformation) when the weight is set to 0, which is precisely what the modified equation achieves. 
As follows, the modified CFG ensures that the gradient $\nabla_{\theta}L_{\text{Diff}}=0$ when the deformation strength is 0, where our equation gives,
$$
\begin{aligned}
\epsilon_\phi(\mathbf{z}_{\mathbf{t}}, y, t) & = \epsilon + \alpha (\hat{\epsilon}_{text} - \hat{\epsilon}_{null}) \\
 & = \epsilon + \alpha (\hat{\epsilon}_{null}-\hat{\epsilon}_{null}) \\
& =\epsilon,
\end{aligned}
$$
and consequently, 
\begin{equation}
\nabla_{\theta}\mathcal{L}_{\mathrm{SDS}}(\mathbf{z}, y, \epsilon, t) = (\epsilon_\phi(\mathbf{z}_{\mathbf{t}}, y, t)-\epsilon) \frac{\partial \mathbf{z}_{\mathbf{t}}}{\partial \theta} = 0
\end{equation}

Injecting attention to this modified CFG score ensures a more stablized control of the deformation strength by getting rid of the bias introduced by $\epsilon_{null}$. The modified CFG score thereby provides guidance that aligns more closely with the intuition of interpolating between the ``identity deformation" and ``deformation conditioned on the target prompt."

We additionally note we can replace $\hat{\epsilon}_{text}$ with $\epsilon$, with minimal compromise in quality because we use a significantly high guidance scale $\alpha$ of 100, following the findings of ~\cite{poole2022dreamfusion}. Such a high guidance scale makes $\hat{\epsilon_{text}}$ relatively insignificant compared to the Classifier Free Guidance term, ensuring minimal change in quality. 

\section{Additional Experiments}

\medskip
\noindent
\textbf{Continuous Concept Space.} Here, we explore a scenario when our users might want to explore a continuous space of generated concepts. While we could run our pipeline multiple times with different relative weights between targets, this could easily become prohibitively expensive if users want a smooth continuity (e.g., generating morphing animations). To address this challenge, we observe that our mesh-based representation provides a dense correspondence map between the source and the deformed shapes. Thus, the user could generate a few sample key frame shapes using our method and smoothly interpolated between them. We show this experiment in Figure \ref{fig:keypoint}

\begin{figure}

\begin{center}
\centering
\begin{overpic}[width=\columnwidth]{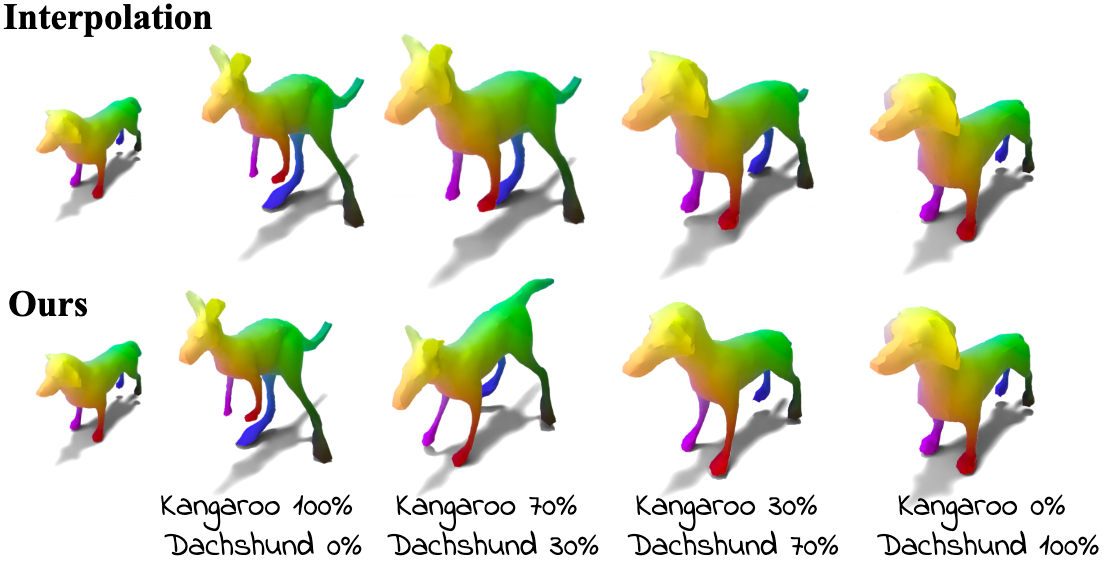}
    \end{overpic}
    \captionof{figure}{{\bf Comparison to naive interpolation.} We compare our two-way blending results (\textit{bottom}) with a naive interpolation of vertices coordinates (\textit{top}). The control weights $w$ are 1.0, 0.7, 0.5, 0.3, 0.1  for each column.}
    \label{fig:shapeinterp}
\end{center}
\end{figure}

\noindent \textbf{3D ROI Ablation.}
We show an ablation of the 3D ROI map $V_R$ for our localized deformation method in Figure~\ref{fig:localdef_ablation}. Due to the viewpoint consistency of our 3D ROI map, our method can generate smooth, meaningful mixing results that respect the specified local regions for each target. In contrast, the results we get without using the 3D ROI map (by directly using the 2D ROI map extracted from self-attention maps), where we observe sharp artifacts as well as significant loss of details for the specified targets. 

\medskip
\noindent \textbf{Modified Classifier Free Guidance Ablation.}
In Figure~\ref{fig:mod_cfg}, we show an ablation of our modified Classifier Free Guidance for self-blending experiments. We notice that the effect of this modification is particularly crucial on smaller weights (typically from 0\% to 15\% of target weight) since for the self-blending application, a low target weight implies more bias from the null text prompt. Notice that the results conditioned on smaller weights without our modified classifier free guidance are severely irregular and biased, while using the modified classifier free guidance stabilizes this bias, regulating the deformation in a much stable, smooth manner.

\vspace{3mm}
\begin{figure*}[tb!]
\begin{center}

\centering
`   \begin{overpic}[width=\linewidth]{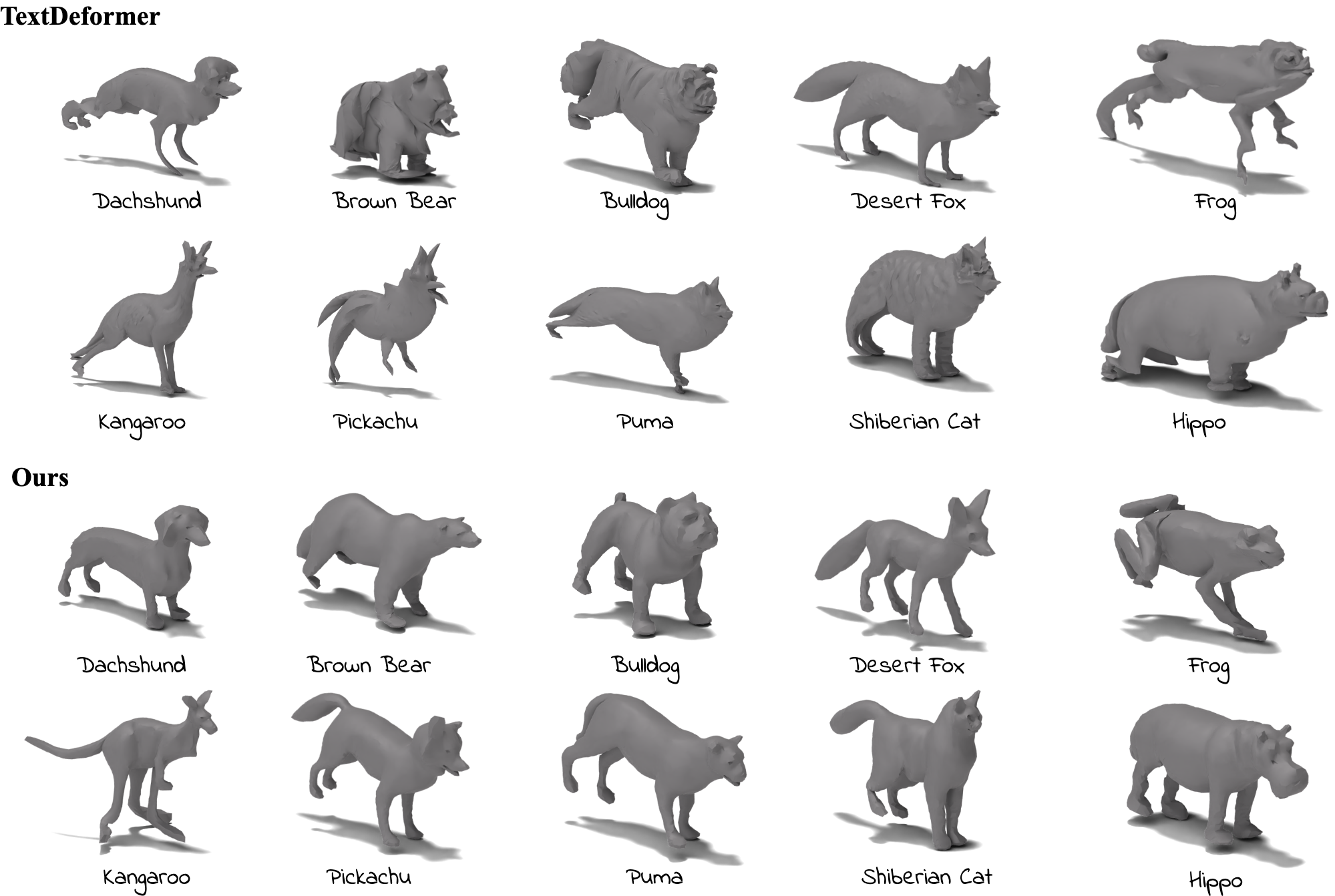}
    \end{overpic}
    \vspace{0.1mm}
    \captionof{figure}{{\bf Comparison of our method and TextDeformer.} \ourmethod{} archives higher detail deformation results with less artifacts. }
    \vspace{-0.3cm}
    \label{fig:textdefcomp}
\end{center}
\end{figure*}

\section{Comparisons}

Since our method is the first to address the concept mixing in mesh deformation, we create simple baselines using existing methods and compare our results to what these baselines can achieve in both qualitative and quantitative metrics. 

\medskip
\noindent\textbf{Shape Interpolation Baseline.}
We consider a simple interpolation baseline, where we generate deformations of two distinct targets using our single target SDS baseline, and directly interpolate the vertex positions between the two meshes. We then compare this naive vertex-wise interpolation baseline to results generated from our BSD method in Figure~\ref{fig:shapeinterp}
We notice that the shape interpolation baseline does not enable new features to emerge for the interpolated shapes, while our method clearly prioritizes the emergence of notable features of each target (notice how the legs of the Kangaroo emerge first while the face of the Dachshund is clearly prioritized, respective of the weights given to each target).

\medskip
\noindent\textbf{Perceptual User Study.}
We present two perceptual user studies to evaluate the overall quality of our results. First, we asked 21 users to compare 5 single-target deformation results by TextDeformer and our method and choose the one that better depicts the input text targets, ``bear,'' ``bulldog,'' ``dachshund,'' ``kangaroo,'' and ``frog'' (see Table~\ref{table:user_quality}). We observe that the users clearly prefer the quality of our method over TextDeformer.

\begin{table}[htb]
\centering
\begin{tabular}{l c c}
\toprule
Target & TextDeformer & \ourmethod{} (ours) \\
\midrule
Bear & 0.048 & \textbf{0.952} \\
Frog & 0.095 & \textbf{0.905} \\
Bulldog &  0.0 & \textbf{1.0} \\
Dachshund & 0.0 & \textbf{1.0} \\
Kangaroo & 0.0 & \textbf{1.0} \\
\bottomrule
\end{tabular}
\caption{{\bf Perceptual user study for quality comparison.} We asked 21 users to compare the quality of our method and TextDeformer. The preference rate for each method is the portion of users who chose the result from one method over the other. The users have a strong preference for our method over the compared one.
}
\label{table:user_quality}
\end{table}

We also asked the same users to evaluate the accuracy of the various BSD weights applied to the blending of two targets, ``Siberian Cat'' and ``Hippo,'' by guessing the correct weights from which the 4 different results were created. Specifically, they were asked to choose from the mixing weight pairs (0.2, 0.8), (0.4, 0.6), (0.6, 0,4), and (0.8, 0.2). Table~\ref{table:user_accuracy} summarizes the results

\begin{table}[htb]
\centering
\begin{tabular}{l c}
\toprule
Targets:  & User's Selection  \\
Siberian Cat \% / Hippo \% & Accuracy \\
\midrule
Siberian Cat 80\% / Hippo 20\% & \textbf{85.7\%} \\
Siberian Cat 60\% / Hippo 40\% &  \textbf{66.7\%} \\
Siberian Cat 40\% / Hippo 60\% & \textbf{85.7\%} \\
Siberian Cat 20\% / Hippo 80\% & \textbf{90.5\%} \\
\bottomrule
\end{tabular}
\caption{{\bf Perceptual user study for the blending weight of our BSD.} We asked 21 users to guess the weights applied to each BSD deformation. The percentages for each section denote the number of users who guessed the weights correctly. In the majority of the blending settings, the users selected the right mixing percentages. This finding suggests that the blending weights properly reflected the level of influence of each target objecting on the resulting deformed shape.
}
\label{table:user_accuracy}
\end{table}

\begin{table}[h]
\centering
\begin{tabular}{@{~}lc@{~~~}c@{~~~}c@{~}}
\toprule
&\multicolumn{3}{c}{CLIP R-Precision $\uparrow$}\\
{\small Method} & {\small CLIP B/14@336px} & {\small CLIP B/16} & {\small CLIP B/32} \\
\midrule
TextDeformer & 0.7  & 0.8 & 0.8 \\
\textbf{Ours} & \textbf{0.8} & 0.8 & \textbf{0.9} \\
\bottomrule
\end{tabular}
\caption{{ \bf Quantitative evaluation.} We compare \ourmethod{} to TextDeformer~\cite{Gao_2023_SIGGRAPH} and report CLIP R-Precision~\cite{CLIP} scores.
%Our results do not perform better for some metric as TextDeformer is directly supervised using CLIP while our method is based on the t5 encoder.
}
\label{table:clip_metrics}
\end{table}

We highlight the significant accuracy of the users' guesses, suggesting that the BSD weights control the concept blending in an intuitively plausible manner. The set of renderings that we use for both evaluations can be found in Figure \ref{fig:textdefcomp}.

\vspace{3mm}

\begingroup
\begin{figure}
\begin{center}
\centering
    \begin{overpic}[width=\columnwidth]{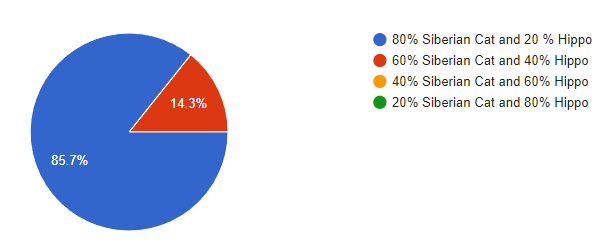}
    \end{overpic}
    \caption{User response for 80\% Siberian cat and 20\% Hippo result.}
    \label{fig: scale_eval_1}
\end{center}
\end{figure}

\begin{figure}
\begin{center}
\centering
    \begin{overpic}[width=\columnwidth]{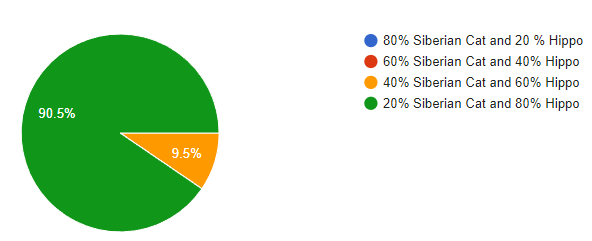}
    \end{overpic}
    \caption{User response for 20\% Siberian cat and 80\% Hippo result.}
    \label{fig: scale_eval_2}
\end{center}
\end{figure}

\begin{figure}
\begin{center}
\centering
    \begin{overpic}[width=\columnwidth]{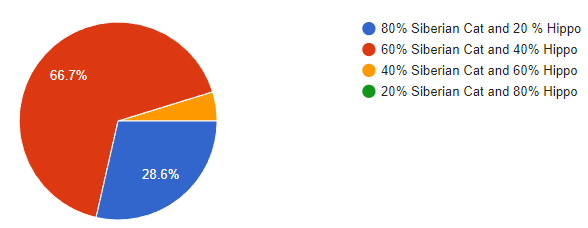}
    \end{overpic}
    \caption{User response for 60\% Siberian cat and 40\% Hippo result.}
    \label{fig: scale_eval_3}
\end{center}
\end{figure}

\begin{figure}
\begin{center}
\centering
    \begin{overpic}[width=\columnwidth]{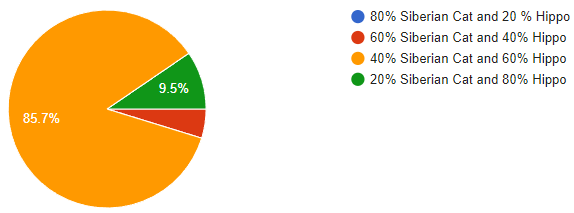}
    \end{overpic}
    \caption{User response for 40\% Siberian cat and 60\% Hippo result.}
    \label{fig: scale_eval_4}
\end{center}
\end{figure}
\endgroup

\noindent\textbf{Quantitative Comparison}
We use the same dataset to quantitatively compare our method to TextDeformer using CLIP R-Precision Scores. We use a source dog mesh and warp it to the following 10 different prompts: ``bear'', ``bulldog'', ``dachshund'', ``desertfox'', ``frog'', ``hippo'', ``kangaroo'', ``pig'', ``puma'', ``siberian cat.'' We evaluate our result using CLIP R-Precision score and show results in Table ~\ref{table:clip_metrics}. Our method outperforms TextDeformer on most metrics, despite the metric being inherently favorable to TextDeformer, since its optimization is directly supervised by CLIP score. 
\end{document}